\documentclass{article}
\usepackage[utf8]{inputenc}
\usepackage[margin=1in]{geometry}

\usepackage{microtype}
\usepackage{graphicx}
\usepackage{subfigure}
\usepackage{booktabs}
\usepackage{natbib}
\setcitestyle{open={(},close={)}}

\usepackage[colorlinks=true,citecolor=blue]{hyperref}

\usepackage{amsmath,amsthm,amsfonts,amssymb,mathdots,array,mathrsfs,bm,bbm,stmaryrd,graphicx,subfigure,xcolor}
\usepackage{algpseudocode,algorithm,algorithmicx}
\usepackage{breakcites}

\usepackage[T1]{fontenc}
\usepackage{enumerate}
\usepackage{inputenc}

\usepackage{graphicx} 
\usepackage{subfigure}

\usepackage{booktabs,balance}
\usepackage{rotating}
\usepackage{boldline}
\usepackage{makecell}
\usepackage{multirow}
\usepackage{balance}

\usepackage{tikz}

\newtheorem{theorem}{Theorem}[section]

\newtheorem{lemma}[theorem]{Lemma}
\newtheorem{proposition}[theorem]{Proposition}

\newtheorem{definition}{Definition}[section]

\newtheorem{remark}{Remark}[section]

\newcommand{\bsmat}{\begin{bmatrix} }
\newcommand{\esmat}{\end{bmatrix} }

\newtheorem{manualtheoreminner}{Theorem}
\newenvironment{manualtheorem}[1]{%
  \renewcommand\themanualtheoreminner{#1}%
  \manualtheoreminner
}{\endmanualtheoreminner}

\newtheorem{manuallemmainner}{Lemma}
\newenvironment{manuallemma}[1]{%
  \renewcommand\themanuallemmainner{#1}%
  \manuallemmainner
}{\endmanuallemmainner}

\newtheorem{manualpropositioninner}{Proposition}
\newenvironment{manualproposition}[1]{%
  \renewcommand\themanualpropositioninner{#1}%
  \manualpropositioninner
}{\endmanualpropositioninner}

\usepackage{environ}
\NewEnviron{smallequation}{%
    \begin{equation}
    \scalebox{0.97}{$\BODY$}
    \end{equation}
    }
    \NewEnviron{smallalign}{%
    \begin{equation}
    \scalebox{0.97}{$\BODY$}
    \end{equation}
    }

\begin{document}

\title{\bf C-MinHash: Rigorously~Reducing~$K$~Permutations~to~Two }

\author{\textbf{Xiaoyun Li and Ping Li} \\\\
Cognitive Computing Lab\\
Baidu Research\\
10900 NE 8th St. Bellevue, WA 98004, USA\\
  \texttt{\{xiaoyunli,\ liping11\}@baidu.com}
}

\date{\vspace{0.5in}}
\maketitle

\begin{abstract}
\noindent\footnote{This work was mostly conducted in 2019 and early 2020, while Xiaoyun Li was still a PhD research intern at Baidu Research. }
Minwise hashing (MinHash) is an important and practical algorithm for generating random hashes to approximate the Jaccard (resemblance) similarity in massive binary (0/1) data. The basic theory of MinHash requires applying hundreds or even thousands of independent random permutations to each data vector in the dataset, in order to obtain reliable results for (e.g.,) building large-scale learning models or approximate near neighbor search in massive data. In this paper, we propose {\bf Circulant MinHash (C-MinHash)} and provide the surprising theoretical results that we just need \textbf{two} independent random permutations.  For C-MinHash, we first conduct an initial permutation on the data vector, then we use a second permutation to generate hash values.  Basically, the second permutation is re-used $K$ times via circulant shifting to produce $K$ hashes. Unlike classical MinHash, these $K$ hashes are obviously correlated,  but we are able to provide rigorous proofs that we still obtain an unbiased estimate of the Jaccard similarity and the theoretical variance is uniformly smaller than that  of the classical MinHash with $K$ independent permutations. The theoretical proofs of C-MinHash require some non-trivial efforts. Numerical experiments are conducted to justify the theory and demonstrate the effectiveness of C-MinHash.

\end{abstract}

\newpage

\section{Introduction} \label{sec:intro}

Given two $D$-dimensional binary vectors $\bm v, \bm w\in\{0,1\}^D$, the Jaccard similarity (also known as the ``resemblance'') is defined as \begin{equation} \label{def:jaccard}
    J(\bm v,\bm w)=\frac{\sum_{i=1}^D \mathbbm 1\{\bm v_i=\bm w_i=1\}}{\sum_{i=1}^D \mathbbm 1\{\bm v_i+\bm w_i\geq 1\}},
\end{equation}
which is a commonly used similarity metric in machine learning and web search applications. The vectors $\bm v$ and $\bm w$ can also be viewed as two sets of items (which represent the locations of non-zero entries), the Jaccard similarity can be equivalently viewed as the size of set intersection over the size of set union. For binary dataset, the Jaccard appears to be a more natural measure of similarity than the ``cosine'', but it is not the focus of this paper to argue which similarity measure should be used for binary data.

The well-known method of ``minwise hashing'' (or MinHash)~\citep{Proc:Broder,Proc:Broder_WWW97,Proc:Broder_STOC98,Proc:Li_Church_EMNLP05,Article:Li_Konig_CACM11} is a standard technique for computing/estimating the Jaccard similarity in massive binary datasets,  with numerous applications such as near neighbor search, duplicate detection, malware detection, web search, clustering, large-scale learning, social networks, computer vision, etc.~\citep{Proc:Charikar_STOC02,Proc:Fetterly_WWW03,Proc:Henzinger_SIGIR06,Proc:Das_WWW07,Proc:Buehrer_WSDM08,Proc:Bendersky_WSDM09,Proc:Chierichetti_KDD09,Proc:Lee_ECCV10,Proc:Li_NIPS11,Proc:Deng_CIKM12,Proc:Chum_CVPR12,Proc:Shrivastava_ECML12,Proc:He_CVPR13,Proc:Tamersoy_KDD14,Proc:Shrivastava_AISTATS14,Article:Zamora16}. The basic idea of MinHash is deceptively simple, as described in Algorithm~\ref{alg:MinHash}.

\subsection{A Review of Minwise Hashing (MinHash)}

\begin{algorithm}[h]{
	\textbf{Input:} Binary data vector $\bm v\in\{0,1\}^D$, \hspace{0.2in}$K$ independent permutations $\pi_1,...,\pi_K$: $[D]\rightarrow[D]$.

\vspace{0.1in}	
	\textbf{Output:} $K$ hash values $h_1(\bm v),...,h_K(\bm v)$.
	
	\vspace{0.05in}
	
	For $k=1$ to $K$
	
	\vspace{0.05in}
	
	\hspace{0.2in}$h_k(\bm v) \leftarrow \min_{i:v_i\neq 0} \pi_k(i)$

	\vspace{0.05in}	
	End For
	\vspace{0.05in}
	}\caption{Minwise-hashing (MinHash) }
	\label{alg:MinHash}
\end{algorithm}

For simplicity, Algorithm~\ref{alg:MinHash} considers just one vector $\bm v\in\{0,1\}^D$. In order to generate $K$ hash values for $\bm v$, we assume $K$ independent permutations:  $\pi_1,...,\pi_K:[D]\mapsto[D]$. For  each permutation, the hash value is the first non-zero location in the permuted vector, i.e.,
\begin{align}\notag
h_k(\bm v)=\min_{i: v_i\neq 0} \pi_k(i),\hspace{0.2in} \forall k=1,...,K.
\end{align}
Similarly, for another binary vector $\bm w\in\{0,1\}^D$, using the same $K$ permutations, we can also obtain $K$ hash values, $h_k(\bm w)$. The estimator of $J(\bm v, \bm w)$, i.e., the Jaccard similarity between $\bm v$ and $\bm w$, is simply
\begin{align}\label{MH-estimator}
    \hat J_{MH}(\bm v,\bm w)=\frac{1}{K}\sum_{k=1}^K \mathbbm 1\{h_k(\bm v)=h_k(\bm w)\},
\end{align}
where $\mathbbm 1\{\cdot\}$ is the indicator function. By fundamental probability and the independence among the permutations, it is easy to show that
\begin{align} \label{MH-var}
    \mathbb E[\hat J_{MH}]=J,\hspace{0.4in} Var[\hat J_{MH}]=\frac{J(1-J)}{K}.
\end{align}

How large is $K$? The answer depends on the application domains. For example, for training large-scale machine learning models, it appears that $K=512$ or $K=1024$ might be sufficient~\citep{Proc:Li_NIPS11}. However, for approximate near neighbor search using many hash tables~\citep{Proc:Indyk_Motwani_STOC98}, it is  likely that $K$ might have to be much larger than $1024$~\citep{Proc:Shrivastava_ECML12,Proc:Shrivastava_AISTATS14}.

In the early work of MinHash~\citep{Proc:Broder,Proc:Broder_WWW97}, actually only one permutation was used by storing the first $K$ non-zero locations after the permutation. Later \cite{Proc:Li_Church_EMNLP05} proposed better estimators to improve the estimation accuracy. The major drawback of the original scheme was that the hashed values did not form a metric space (e.g., satisfying the triangle inequality) and hence could not be used in numerous algorithms/applications which require metric space. We believe this was the main reason why the original authors moved to using $K$ permutations~\citep{Proc:Broder_STOC98}.

\subsection{From $K$ Permutations to 2 Permutations}

In this paper, we present some (perhaps surprising) theoretical findings that we just need 2 permutations for MinHash and the results (estimation variances) are even more accurate. Basically, with the \textbf{initial permutation} (denoted by $\sigma$), we randomly shuffle the data to break whatever structure which might exist in the original data, and then the \textbf{second permutation} (denoted by $\pi$) is applied and re-used $K$ times to generate $K$ hash values, via a simple ``circulant'' trick. Therefore, we name the proposed method \textbf{C-MinHash}, i.e., circulant MinHash.

The ``circulant'' trick was used in the literature of random projections. For example, \cite{Article:Yu_JMLR17} showed that using the circulant trick, the estimation accuracy of random projections was hurt, but not by too much when the data are sparse. In this paper, we show the surprising theoretical results, in Theorem~\ref{theo:CMH-sigma,pi var} and Theorem~\ref{theo:smaller-variance}, that C-MinHash actually exhibits strictly smaller variances than MinHash. While the proofs require some non-trivial efforts, the correctness of the theorems can be easily verified by simulations. \\

\noindent\textbf{Roadmap:} \ In this paper, we will present two variants of C-MinHash. In Section~\ref{sec:CMH,0,pi}, the initial permutation $\sigma$ is actually not used and we directly use permutation $\pi$ on the original data vector to generate $K$ hashes. We name this method ``C-MinHash-$(0,\pi)$''. Although it is not our recommended method, our  analysis for C-MinHash-$(0,\pi)$ provides the necessary preparation for later methods and the intuition for understanding the need for the initial permutation. In Section~\ref{sec:CMH,sigma,pi}, we analyze the recommended method ``C-MinHash-$(\sigma,\pi)$'', i.e., we use both the initial permutation $\sigma$ and the second permutation $\pi$. The theoretical results demonstrate that the variance of C-MinHash-$(\sigma,\pi)$ is uniformly smaller than that of the original MinHash. Section~\ref{sec:experiment} provides the experiments to sanity check our theoretical findings.

\vspace{0.1in}

\section{C-MinHash-$(0,\pi)$: Circulant MinHash without the Initial\\ Permutation} \label{sec:CMH,0,pi}

\begin{algorithm}[h]
{
	\textbf{Input:} Binary data vector $\bm v\in\{0,1\}^D$, \hspace{0.15in} Permutation vector $\pi$: $[D]\rightarrow[D]$
	
	\vspace{0.05in}
	
	\textbf{Output:} Hash values $h_1(\bm v),...,h_K(\bm v)$
	
	\vspace{0.1in}

	For $k=1$ to $K$
	
	\vspace{0.05in}
	
	\hspace{0.2in}Shift $\pi$ circulantly rightwards by $k$ units: $\pi_k=\pi_{\rightarrow k}$
	
	\vspace{0.05in}
	
	\hspace{0.2in}$h_k(\bm v)\leftarrow \min_{i:v_i\neq 0} \pi_{\rightarrow k}(i)$

	\vspace{0.05in}
	
	End For

	}\caption{C-MinHash-$(0,\pi)$}
	\label{alg:C-MinHash}
\end{algorithm}

As shown in Algorithm~\ref{alg:C-MinHash}, the C-MinHash algorithm has similar operations as MinHash. The difference lies in the permutations used in the hashing process. To generate each hash $h_k(\bm v)$, we permute the data vector using $\pi_{\rightarrow k}$, which is the permutation shifted $k$ units circulantly towards right based on $\pi$. For example,
\begin{align*}
\pi=[3,1,2,4],\hspace{0.2in} \pi_{\rightarrow 1}=[4,3,1,2],\hspace{0.2in}  \pi_{\rightarrow 2}=[2,4,3,1].
\end{align*}

\newpage

\begin{figure}[t]

\centering
\includegraphics[width=4.5in]{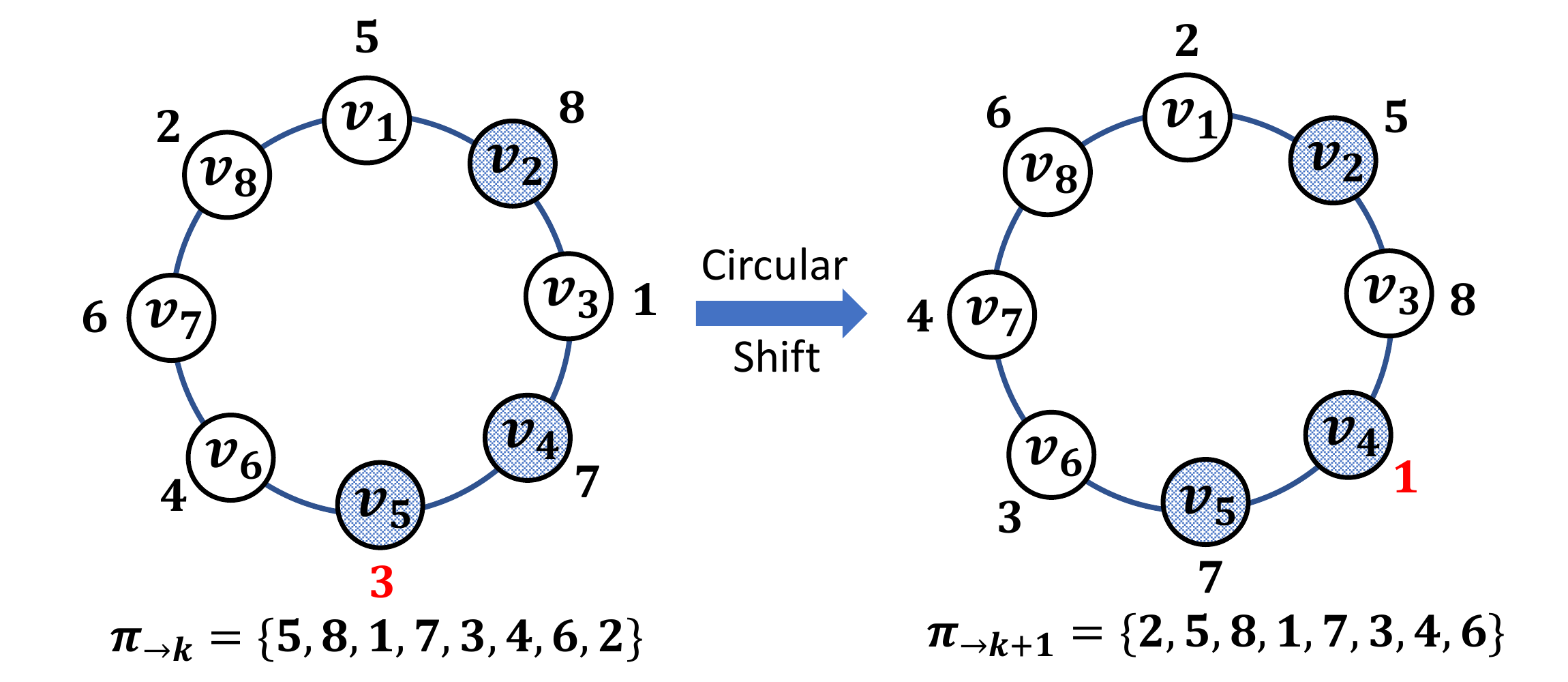}

\vspace{-0.1in}
\caption{An illustration of the idea of C-MinHash. The data vector has three non-zeros, $v_2=v_4=v_5=1$. In this example, $h_k(\bm v)=3$, $h_{k+1}(\bm v)=1$.}
\label{fig:scheme}\vspace{0in}
\end{figure}

Conceptually, we may think of circulation as concatenating the first and last elements of a vector to form a circle; see Figure~\ref{fig:scheme}. We  set the hash value $h_k(\bm v)$ as the position of the first non-zero after being~permuted~by~$\pi_{\rightarrow k}$.  Analogously, we define the C-MinHash-$(0,\pi)$ estimator of the Jaccard similarity $J(\bm v, \bm w)$ as
\begin{equation}
    \hat J_{0,\pi}=\frac{1}{K}\sum_{k=1}^K \mathbbm 1\{h_k(\bm v)=h_k(\bm w)\}, \label{eqn:CMH0pi estimator}
\end{equation}
where $h$ is the hash value output by Algorithm~\ref{alg:C-MinHash}. In this paper, for simplicity, we assume $K\leq D$. \\

Next, we  present the theoretical analysis for Algorithm~\ref{alg:C-MinHash}, in terms of the expectation (mean) and the variance of the estimator $\hat J_{0,\pi}$. Our results reveal that the estimation accuracy depends on the initial data distribution, which may lead to undesirable performance behaviors when  real-world datasets exhibit various structures.  On the other hand, while it is not our recommended method, the analysis of C-MinHash-$(0,\pi)$ serves a good preparation (and insight) for the analysis of C-MinHash-$(\sigma,\pi)$ which will soon be described. \\

Here we introduce some notations and definitions, before we proceed with the theoretical analysis. Firstly, given $\bm v,\bm w\in \{0,1\}^D$,  we define $a$ and $f$ as follows:
\begin{align}  \label{def:a, f}
a = \sum_{i=1}^D\mathbbm 1\{ \bm v_i = 1 \text{ and }  \bm w_i = 1\},\hspace{0.2in}
f = \sum_{i=1}^D\mathbbm 1\{ \bm v_i = 1 \text{ or }  \bm w_i = 1\}.
\end{align}
We say that $(\bm v,\bm w)$ is a \textit{$(D,f,a)$-data pair}, whose Jaccard similarity can also be written as $J = a/f$.

\begin{definition}\label{def-1}
Consider two binary vectors $\bm v,\bm w\in \{0,1\}^D$. Define the \textbf{location vector} as $\bm x\in\{O,\times,-\}^D$, with $\bm x_i$ being ``$O$'', ``$\times$'', ``$-$'', when $\bm v_i=\bm w_i=1$, $\bm v_i+\bm w_i=1$ and $\bm v_i=\bm w_i=0$, respectively.
\end{definition}

The location vector $\bm x$ can fully characterize a hash collision. When a permutation $\pi_{\rightarrow k}$ is applied, the hash values $h_k(\bm v)$ and $h_k(\bm w)$ would collide if after permutation, the first ``$O$'' is placed before the first ``$\times$'' (counting from small to large). This observation will be the key in our theoretical analysis.

\begin{definition}\label{def-sets}
For $A,B\in\{O,\times,-\}$, let $\{(A,B)|\triangle\}$ denote the set $\{(i,j):(\bm x_i,\bm x_j)=(A,B),j-i=\triangle\}$. For each $1\leq \triangle\leq K-1$, define
\begin{align*}
    &\mathcal L_0(\triangle)=\{(O,O)|\triangle\},\hspace{0.05in}\mathcal L_1(\triangle)=\{(O,\times)\},\hspace{0.05in}\mathcal L_2(\triangle)=\{(O,-)\},\\
    &\mathcal G_0(\triangle)=\{(-,O)|\triangle\},\hspace{0.05in}\mathcal G_1(\triangle)=\{(-,\times)\}\hspace{0.05in}, \mathcal G_2(\triangle)=\{(-,-)\},\\
    &\mathcal H_0(\triangle)=\{(\times ,O)|\triangle\},\hspace{0.05in}\mathcal H_1(\triangle)=\{(\times ,\times)\},\hspace{0.05in} \mathcal H_2(\triangle)=\{(\times ,-)\}.
\end{align*}
\end{definition}
\begin{remark}
For the ease of notation, by circulation we write $\bm x_j=\bm x_{j-D}$ when $D<j<2D$.
\end{remark}

Definition~\ref{def-sets} measures the relative location of different types of points in the location vector, for a specific pair of data vectors. One can easily verify that given fixed $a,f,D$, it holds that, for $\forall 1\leq \triangle\leq K-1$,
\begin{align}
\begin{aligned}
    &|\mathcal L_0|+|\mathcal L_1|+|\mathcal L_2|=|\mathcal L_0|+|\mathcal G_0|+|\mathcal H_0|=a,\\
    &|\mathcal G_0|+|\mathcal G_1|+|\mathcal G_2|=|\mathcal L_2|+|\mathcal G_2|+|\mathcal H_2|=D-f,\\
    &|\mathcal H_0|+|\mathcal H_1|+|\mathcal H_2|=|\mathcal L_1|+|\mathcal G_1|+|\mathcal H_1|=f-a,
\end{aligned} \label{eqn:constraint}
\end{align}
which is the intrinsic constraints on the size of above sets.\\

We are now ready to analyze the expectation and variance of the estimator $\hat J_{0,\pi}$. It is easy to see that $\hat J_{0,\pi}$ is still unbiased, i.e., $\mathbb E[\hat J_{0,\pi}]=J$, by linearity of expectation.  Lemma~\ref{lemma1} provides an important quantity that leads to $Var[\hat J_{0,\pi}]$ as in Theorem~\ref{theo:CMH-0,pi var}.  The proofs of Lemma~\ref{lemma1} and Theorem~\ref{theo:CMH-0,pi var} are given in Appendix~\ref{sec:lemma1 proof} and Appendix~\ref{sec:theo:CMH-0,pi var proof}, respectively.\\

\begin{lemma} \label{lemma1}
 For any $1\leq s<t\leq K$ with $t-s=\triangle$, we have
\begin{equation*}
    \mathbb E_\pi\big[\mathbbm 1\{h_s(\bm v)=h_s(\bm w)\} \mathbbm 1\{h_t(\bm v)=h_t(\bm w)\}\big]=\frac{|\mathcal L_0(\triangle)|+(|\mathcal G_0(\triangle)|+|\mathcal L_2(\triangle)|)J}{f+|\mathcal G_0(\triangle)|+|\mathcal G_1(\triangle)|},
\end{equation*}
where the sets are defined in Definition~\ref{def-sets} and $h_s$, $h_t$ are the hash values as in Algorithm~\ref{alg:C-MinHash}.
\end{lemma}

\begin{theorem} \label{theo:CMH-0,pi var}
Under the same setting as in Lemma~\ref{lemma1}, the variance of $\hat J_{0,\pi}$~is
\begin{align*}
    Var[\hat J_{0,\pi}]=\frac{J}{K}+\frac{2\sum_{s=2}^{K}(s-1)\Theta_{K-s+1}}{K^2}-J^2,
\end{align*}
where $\Theta_\triangle \triangleq E_\pi\big[\mathbbm 1\{h_s(\bm v)=h_s(\bm w)\} \mathbbm 1\{h_t(\bm v)=h_t(\bm w)\}\big]$ as in Lemma~\ref{lemma1} with any $t-s=\triangle$.
\end{theorem}

From Theorem~\ref{theo:CMH-0,pi var}, we see that the variance of $\hat J_{0,\pi}$  depends on $a$, $f$, and the size of sets $\mathcal L$'s and $\mathcal G$'s as in Definition~\ref{def-1}, which is determined by the location vector $\bm x$. Since we use the original data vectors without randomly permuting the entries beforehand, $Var[\hat J_{0,\pi}]$ is called ``\textit{location-dependent}''  as it is dependent on the location of non-zero entries of the original data. 

\section{C-MinHash-$(\sigma,\pi)$: Circulant MinHash with the Independent\\ Initial Permutation}  \label{sec:CMH,sigma,pi}

\begin{algorithm}[h]
{
	\textbf{Input:} Binary data vector $\bm v\in\{0,1\}^D$, \hspace{0.15in} Permutation vectors $\pi$ and $\sigma$: $[D]\rightarrow[D]$
	
	\vspace{0.05in}
	
	\textbf{Output:} Hash values $h_1(\bm v),...,h_K(\bm v)$
	
	\vspace{0.1in}

    Initial permutation: $\bm v'$ = $\sigma(\bm v)$
	
	\vspace{0.05in}
	
	For $k=1$ to $K$
	
	\vspace{0.05in}
	
	\hspace{0.2in}Shift $\pi$ circulantly rightwards by $k$ units: $\pi_k=\pi_{\rightarrow k}$
	
	\vspace{0.05in}
	
	\hspace{0.2in}$h_k(\bm v)\leftarrow \min_{i:v_i'\neq 0} \pi_{\rightarrow k}(i)$

	\vspace{0.05in}
	
	End For
	\vspace{0.05in}

	}\caption{C-MinHash-$(\sigma,\pi)$}
	\label{alg:C-MinHash initial}
\end{algorithm}

The method C-MinHash-$(\sigma,\pi)$ is summarized in Algorithm~\ref{alg:C-MinHash initial}, which is very similar to Algorithm~\ref{alg:C-MinHash} for C-MinHash-$(0,\pi)$. This time we apply an initial permutation $\sigma$ on the data to break whatever structures which might exist.  Similarly, we define the C-MinHash-$(\sigma,\pi)$ estimator of $J$ as
\begin{equation}
    \hat J_{\sigma,\pi}=\frac{1}{K}\sum_{k=1}^K \mathbbm 1\{h_k(\bm v)=h_k(\bm w)\}, \label{eqn:CMHsigmapi estimator}
\end{equation}
where $h_k$'s are the hash values output by Algorithm~\ref{alg:C-MinHash initial}. Again, for simplicity we assume $K\leq D$. In the remaining part of this section, we will present our main theoretical result  in Theorem~\ref{theo:smaller-variance}, that our C-MinHash-$(\sigma,\pi)$ achieves a uniformly smaller estimation variance than that of the classical MinHash.

\vspace{0.1in}

First, by  linearity of expectation and the fact that $\sigma$ and $\pi$ are independent, it is easy to verify that $\hat J_{\sigma,\pi}$ is still an unbiased estimator of  $J$. The following Theorem provides the variance of $\hat J_{\sigma,\pi}$, whose proof can be found in Appendix~\ref{sec:theo:CMH-sigma,pi var proof}.

\begin{theorem} \label{theo:CMH-sigma,pi var}
Let $a,f$ be defined as in (\ref{def:a, f}). When $0<a<f\leq D$ ($J\notin \{0,1\}$), we have
\begin{align}
    Var[\hat J_{\sigma,\pi}]=\frac{J}{K}+\frac{(K-1)\tilde{\mathcal E}}{K}-J^2, \label{eqn:MH_sigma-var}
\end{align}
where with $l=\max(0,D-2f+a)$, and
\begin{align}\notag
    \tilde{\mathcal E}=\sum_{\{l_0,l_2,g_0,g_1\}}\Bigg\{&\left(\frac{l_0}{f+g_0+g_1}+\frac{a(g_0+l_2)}{(f+g_0+g_1)f}\right) \\\label{eqn:E} &\times\sum_{s=l}^{D-f-1}\frac{\binom{D-f}{s}}{\binom{D-a-1}{D-f-1}} \frac{\binom{f-a-1}{D-f-s-1}\binom{s}{n_1}\binom{D-f-s}{n_2}\binom{D-f-s}{n_3}\binom{f-a-(D-f-s)}{n_4}\binom{a-1}{a-l_1-l_2}}{\binom{D-1}{a}}\Bigg\}.
\end{align}
The feasible set $\{l_0,l_2,g_0,g_1\}$ satisfies the intrinsic constraints (\ref{eqn:constraint}), and
\begin{align*}
    &n_1=g_0-(D-f-s-g_1), \hspace{0.33in} n_2=D-f-s-g_1,\\
    &n_3=l_2-g_0+(D-f-s-g_1), \hspace{0.08in} n_4=l_1-(D-f-s-g_1).
\end{align*}
Also, when $a=0$ or $f=a$ ($J=0$ or $J=1$), we have $Var[\hat J_{\sigma,\pi}]=0$.
\end{theorem}

As expected, since the original locational structure of the data is broken by the initial permutation $\sigma$, $Var[\hat J_{\sigma,\pi}]$ only depends on  ($D,f,a$). In the sequel, we investigate the statistical properties of $Var[\hat J_{\sigma,\pi}]$ in detail and present our main result. \\

\begin{figure}[b!]


\begin{center}
		\includegraphics[width=2.7in]{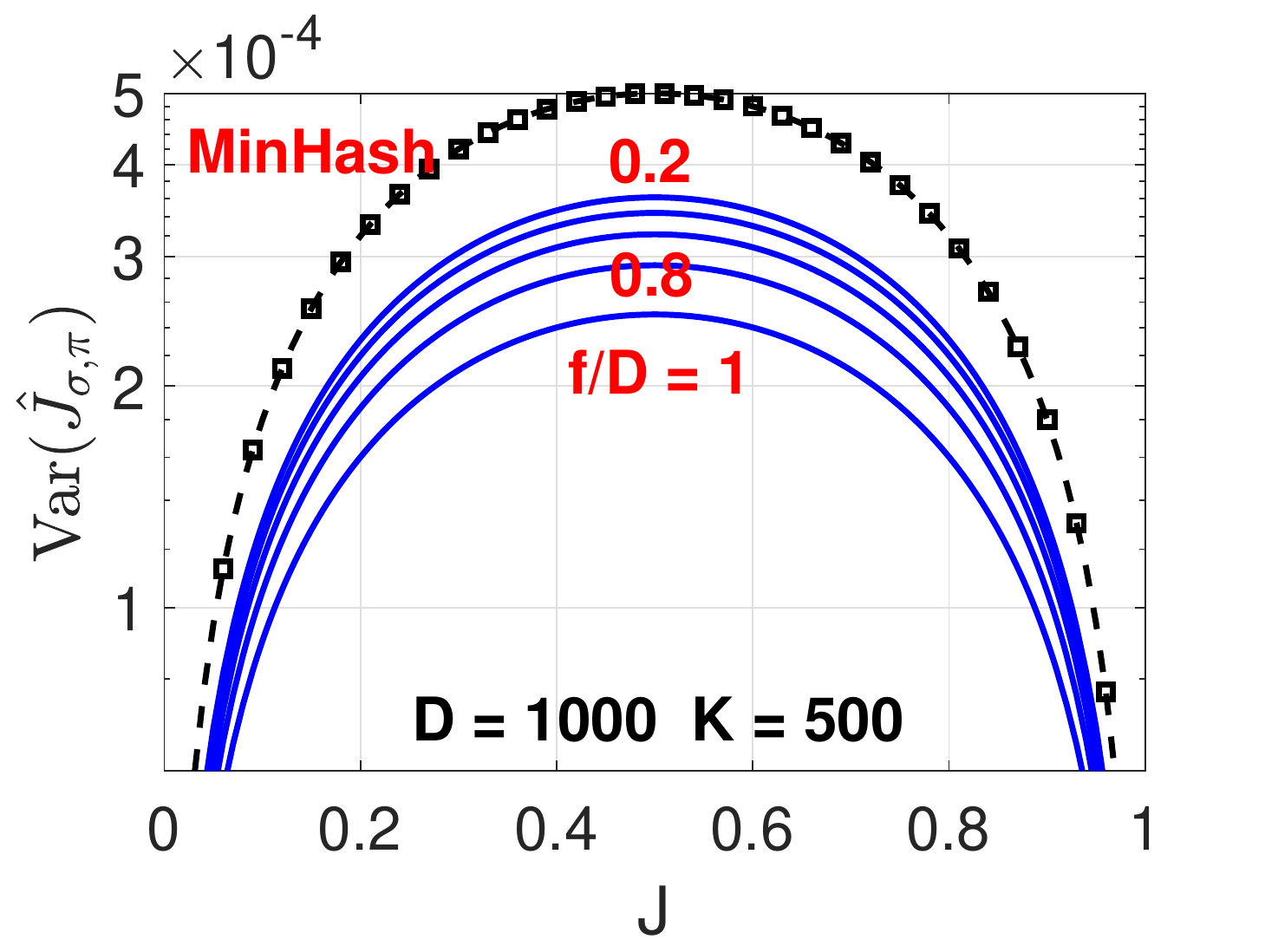}
		\includegraphics[width=2.7in]{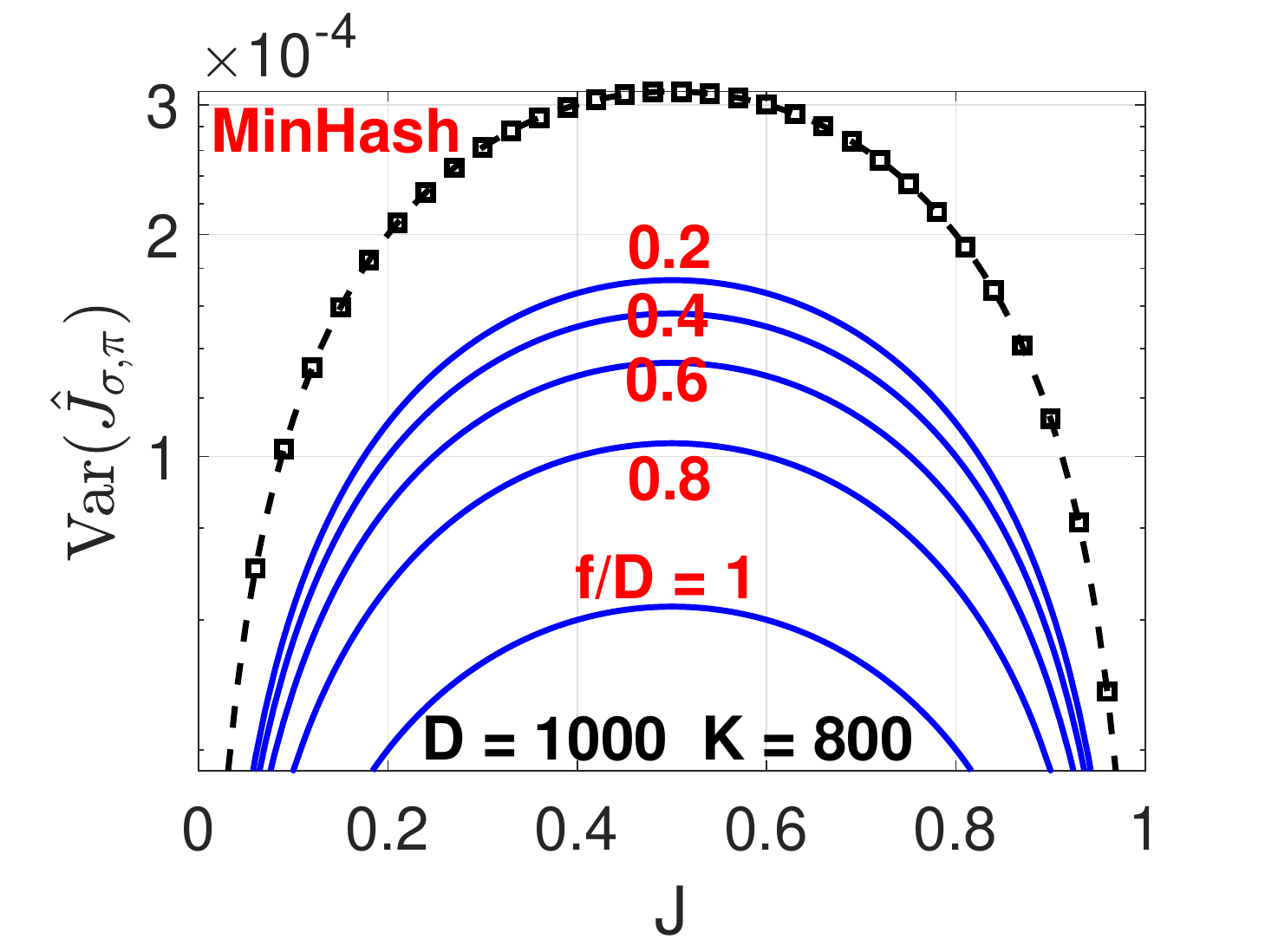}
\end{center}	

	\vspace{-0.15in}

	\caption{$Var[\hat J_{\sigma,\pi}]$ versus $J$, with $D=1000$ and varying $f$, for $K=500$ (left) and $K=800$ (right). We see that $Var[\hat J_{\sigma,\pi}]$ is symmetric about 0.5 (Proposition~\ref{prop:symmetry}) and always smaller than $Var[\hat J_{MH}]$ (Theorem~\ref{theo:smaller-variance}).}
	\label{fig:var-symmetry}
\end{figure}

Firstly, same as MinHash, Proposition~\ref{prop:symmetry}  states that given $D$ and $f$, the variance of $\hat J_{\sigma,\pi}$ is symmetric about $J=0.5$, as illustrated in Figure~\ref{fig:var-symmetry}, which also confirms that the variance of $\hat{J}_{\sigma,\pi}$ is smaller than the variance of the original MinHash. The proof is provided in Appendix~\ref{sec:prop:symmetry proof}.

\begin{proposition}[Symmetry] \label{prop:symmetry}
$Var[\hat J_{\sigma,\pi}]$ is the same for the $(D,f,a)$-data pair and the $(D,f,f-a)$-data pair, $\forall 0\leq a\leq f\leq D$.
\end{proposition}

\begin{figure}[t]
\begin{center}
	\includegraphics[width=2.7in]{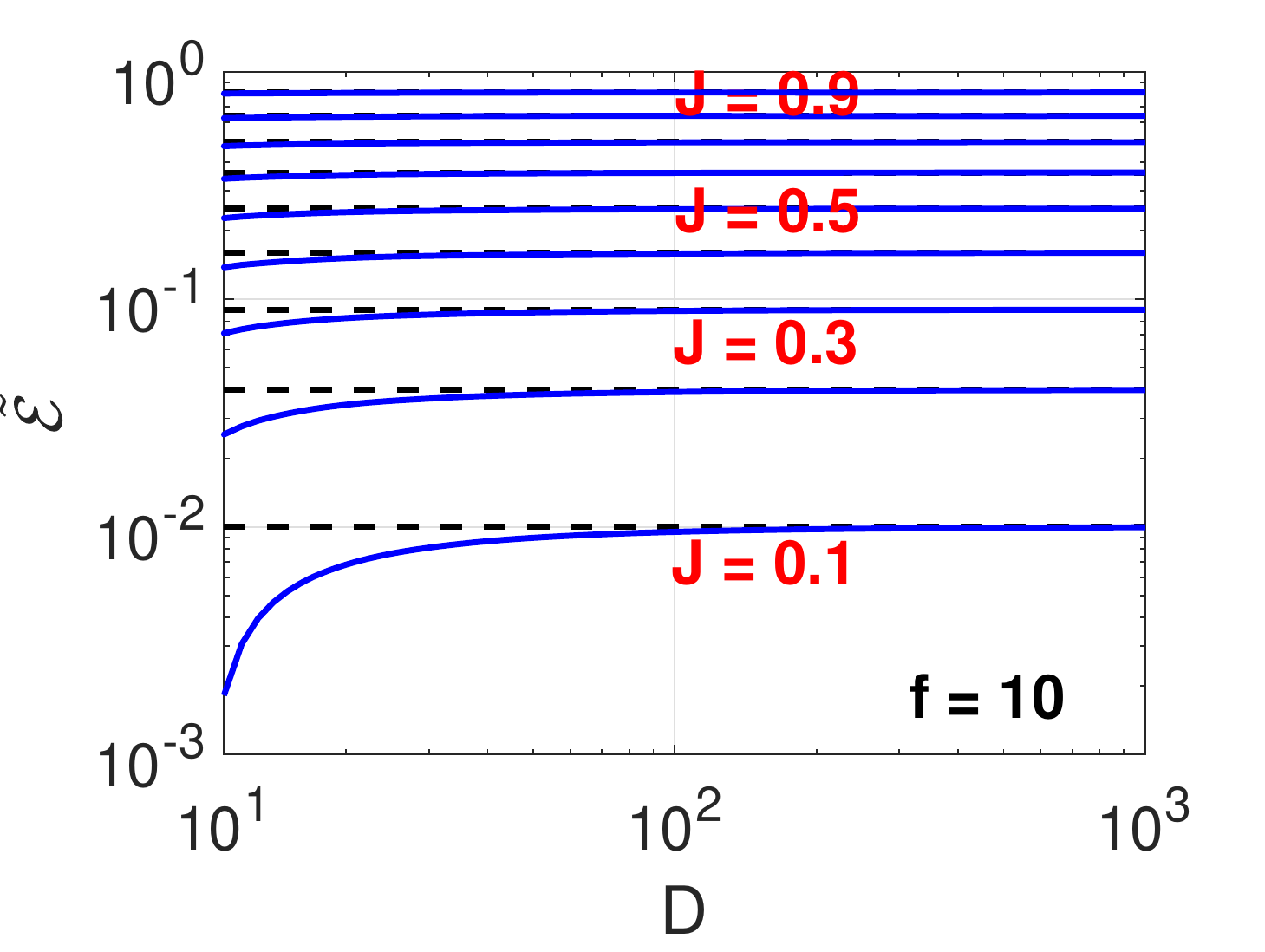}
	\includegraphics[width=2.7in]{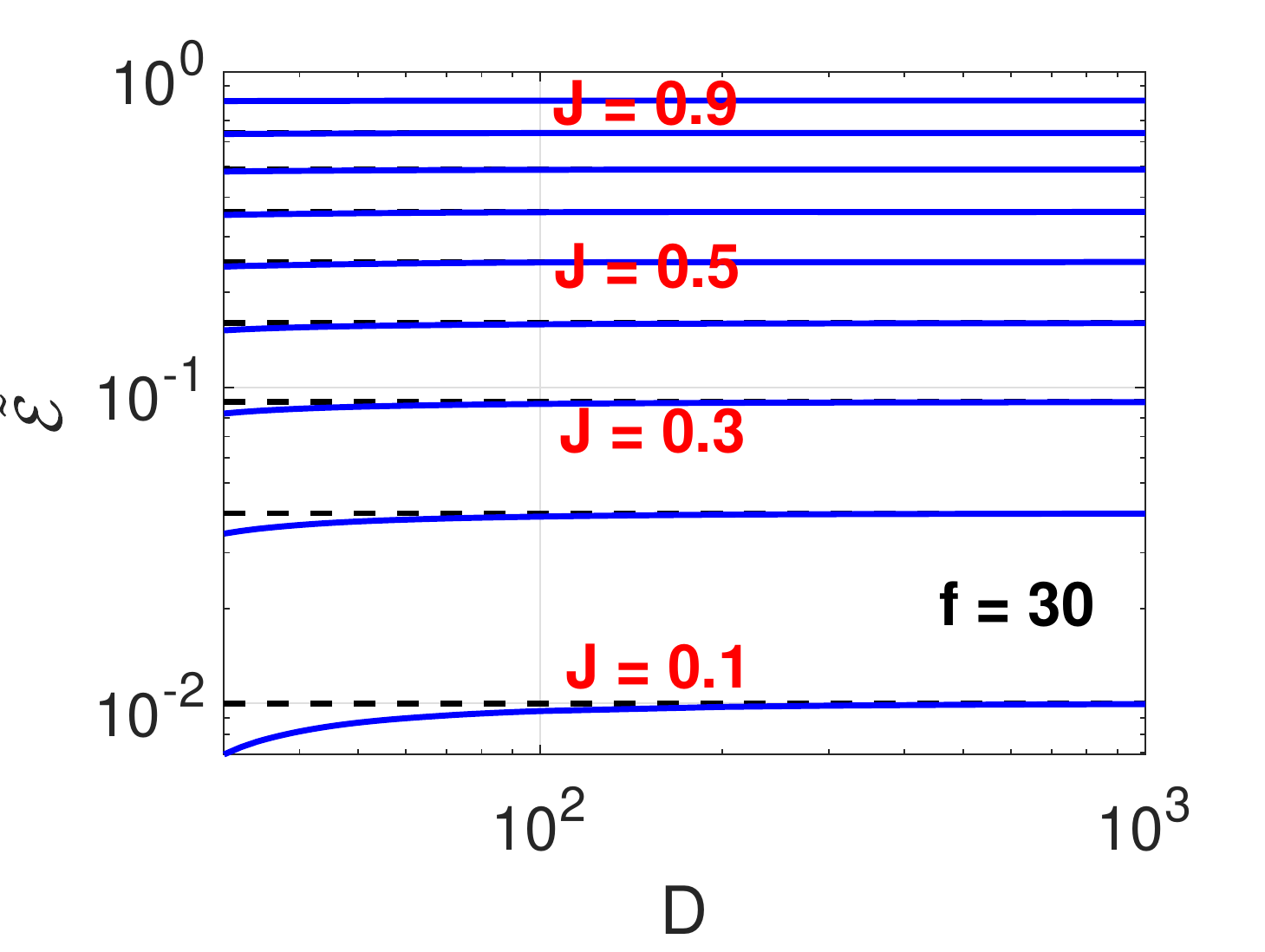}
	\end{center}
	\vspace{-0.2in}
	\caption{Theoretical $\tilde{\mathcal E}$, for $f=10$ (left) and  $f=30$ (right).  Each dash line represents the corresponding $J^2$. We see that $\tilde{\mathcal E}$ is increasing with $D$ and converges to $J^2$, which validates our theory. }
	\label{fig:E tilde}
\end{figure}

A rigorous comparison of  $Var[\hat J_{\sigma,\pi}]$ and $Var[\hat J_{MH}]$ appears to be a  challenging task given the complicated combinatorial form of $Var[\hat J_{\sigma,\pi}]$. The following lemma  characterizes an important property of $\tilde{\mathcal E}$ in \eqref{eqn:E} in  Theorem~\ref{theo:CMH-sigma,pi var}, stating that it is monotone in $D$ when both $a$ and $f$ are fixed, as illustrated in Figure~\ref{fig:E tilde}.

\begin{lemma}[Increasing Increment] \label{lemma:increment}
Assume $f>a>0$  are arbitrary and fixed. Denote $\tilde{\mathcal E}_D$ as in \eqref{eqn:E} in Theorem~\ref{theo:CMH-sigma,pi var}, with $D$ treated as a parameter. Then we have $\tilde{\mathcal E}_{D+1}> \tilde{\mathcal E}_D$ for $\forall D\geq f$.
\end{lemma}

\vspace{0.1in}

Equipped with Lemma~\ref{lemma:increment}, we arrive at the following main theoretical result of this work, on the uniform variance reduction of C-MinHash-$(\sigma,\pi)$.

\begin{theorem}[Uniform Superiority] \label{theo:smaller-variance}
For any two binary vectors $\bm v,\bm w\in\{0,1\}^D$ with $J\neq 0$ or $1$, it holds that $Var[\hat J_{\sigma,\pi}(\bm v,\bm w)]< Var[\hat J_{MH}(\bm v,\bm w)]$.
\end{theorem}

\begin{figure}[h]
\vspace{-0.1in}

\begin{center}
		\includegraphics[width=2.7in]{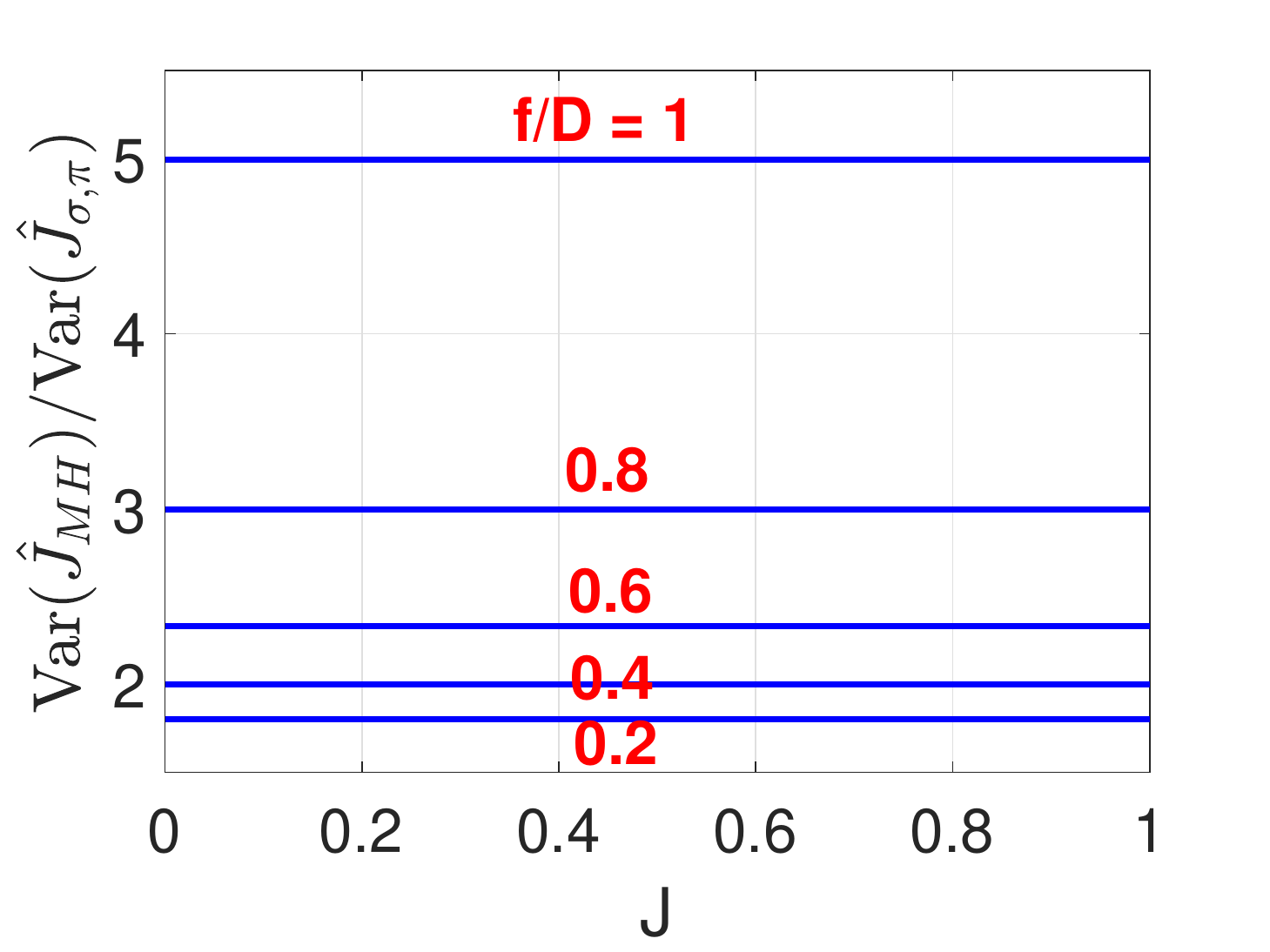}
\end{center}
	
	\vspace{-0.25in}
	
	\caption{Variance ratio $\frac{Var[\hat J_{MH}]}{Var[\hat J_{\sigma,\pi}]}$, for $D=1000$ and $K=800$, confirming Proposition~\ref{prop:same-ratio}. }
	\label{fig:same-ratio}
\end{figure}

Given the uniform superiority, an interesting question is whether C-MinHash-$(\sigma,\pi)$ is more beneficial in the high similarity region. Interestingly, in Figure~\ref{fig:same-ratio} and Theorem~\ref{prop:same-ratio},  we show that the improvement of C-MinHash-$(\sigma,\pi)$ compared with MinHash is actually same for any $J$, for given  $D,f$ and $K$.

\vspace{0.05in}
\begin{proposition}[Consistent Improvement] \label{prop:same-ratio}
Suppose $f$ is fixed. In terms of  $a$, the variance ratio $\frac{Var[\hat J_{MH}(\bm v,\bm w)]}{Var[\hat J_{\sigma,\pi}(\bm v,\bm w)]}$ is constant for any $0<a<f$.
\end{proposition}

\begin{figure}[t]
\begin{center}
		\includegraphics[width=2.7in]{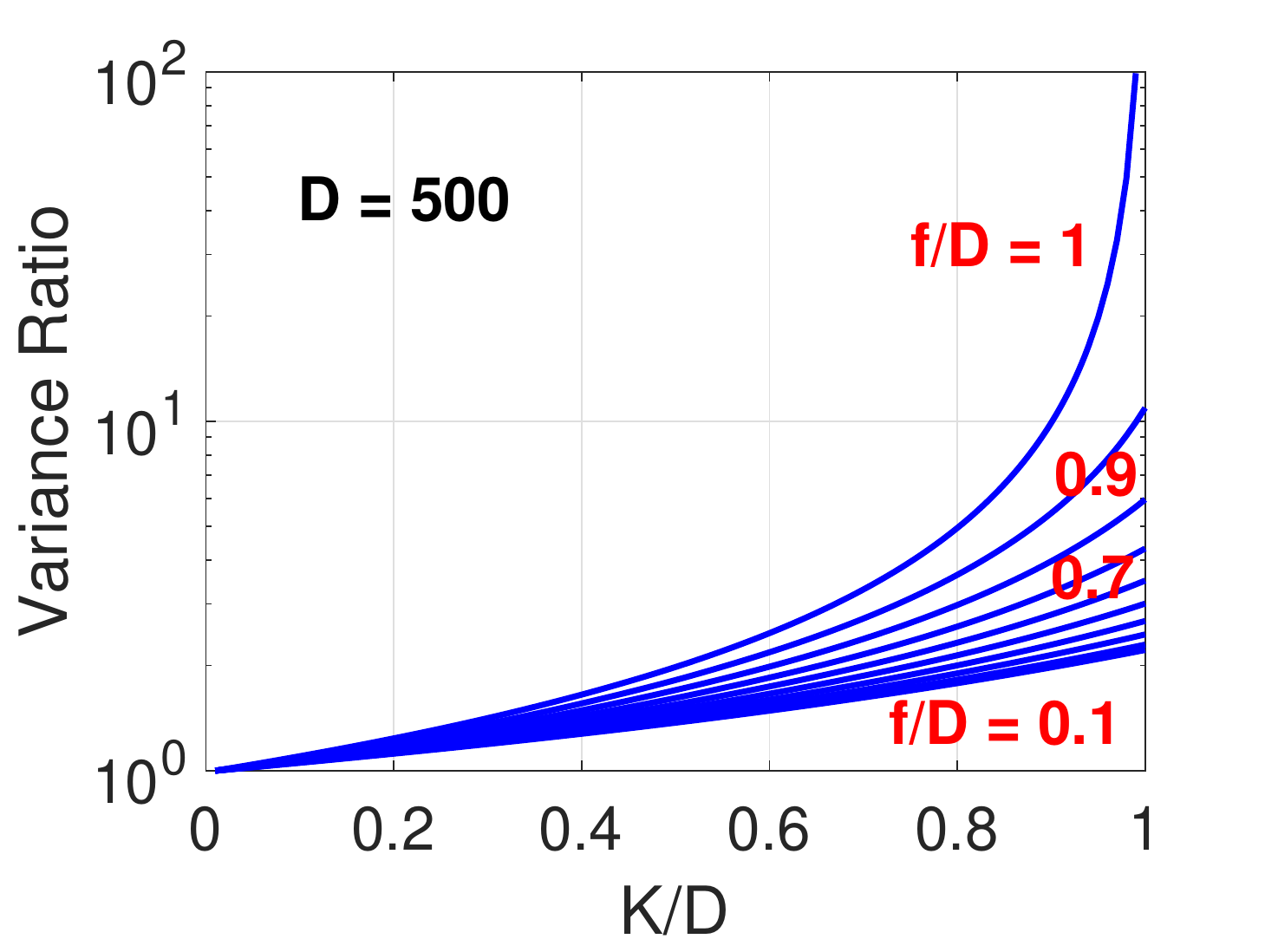}
		\includegraphics[width=2.7in]{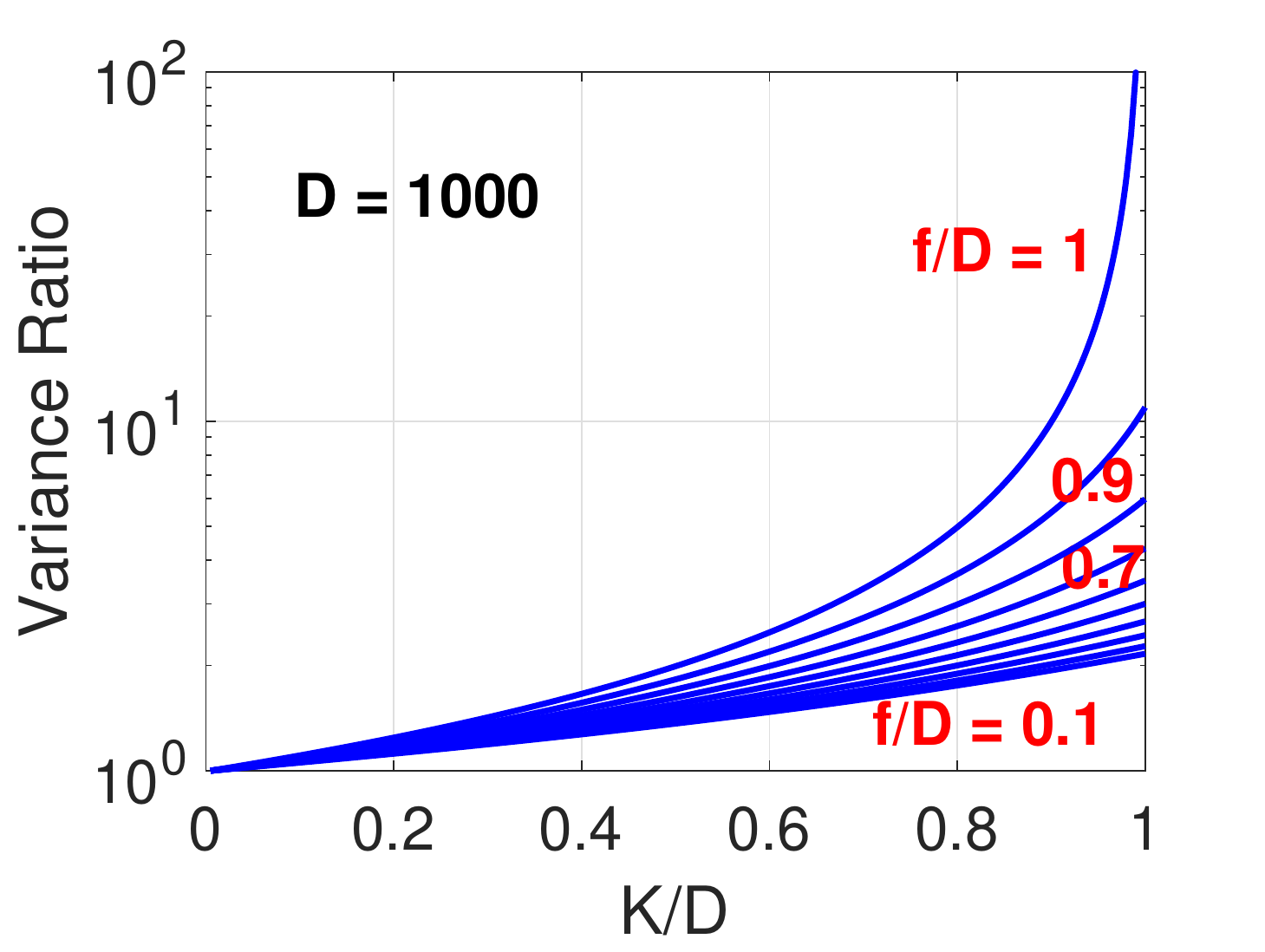}
\end{center}
	\vspace{-0.1in}
	\caption{Variance ratio $\frac{Var[\hat J_{MH}]}{Var[\hat J_{\sigma,\pi}]}$, for $D=500$ (left) and $D=1000$ (right). }
	\label{fig:var-ratio}
\end{figure}

How is the improvement affected by the sparsity (i.e., $f$) and the number of hashes $K$? In Figure~\ref{fig:var-ratio}, we plot the variance ratio $\frac{Var[\hat J_{MH}]}{Var[\hat J_{\sigma,\pi}]}$ with different combinations $f$ and $K$, given fixed $D$. Note that, by Proposition~\ref{prop:same-ratio}, we do not need to consider $a$ here since the variance ratio is independent of $a$. The results in Figure~\ref{fig:var-ratio} once again verify Theorem~\ref{theo:smaller-variance}, i.e., the variance ratio is always greater than 1. Specifically, we see that the improvement in variance increases with $K$ (more hashes) and $f$ (more non-zero entries).

\section{Numerical Experiments}  \label{sec:experiment}

In this section, we provide numerical experiments on synthetic as well as real-world data to validate our theoretical findings and demonstrate that C-MinHash can indeed lead to smaller Jaccard estimation errors.

\subsection{Sanity Check: a Simulation Study}

\begin{figure}
	\begin{center}
		\mbox{
		\includegraphics[width=2.7in]{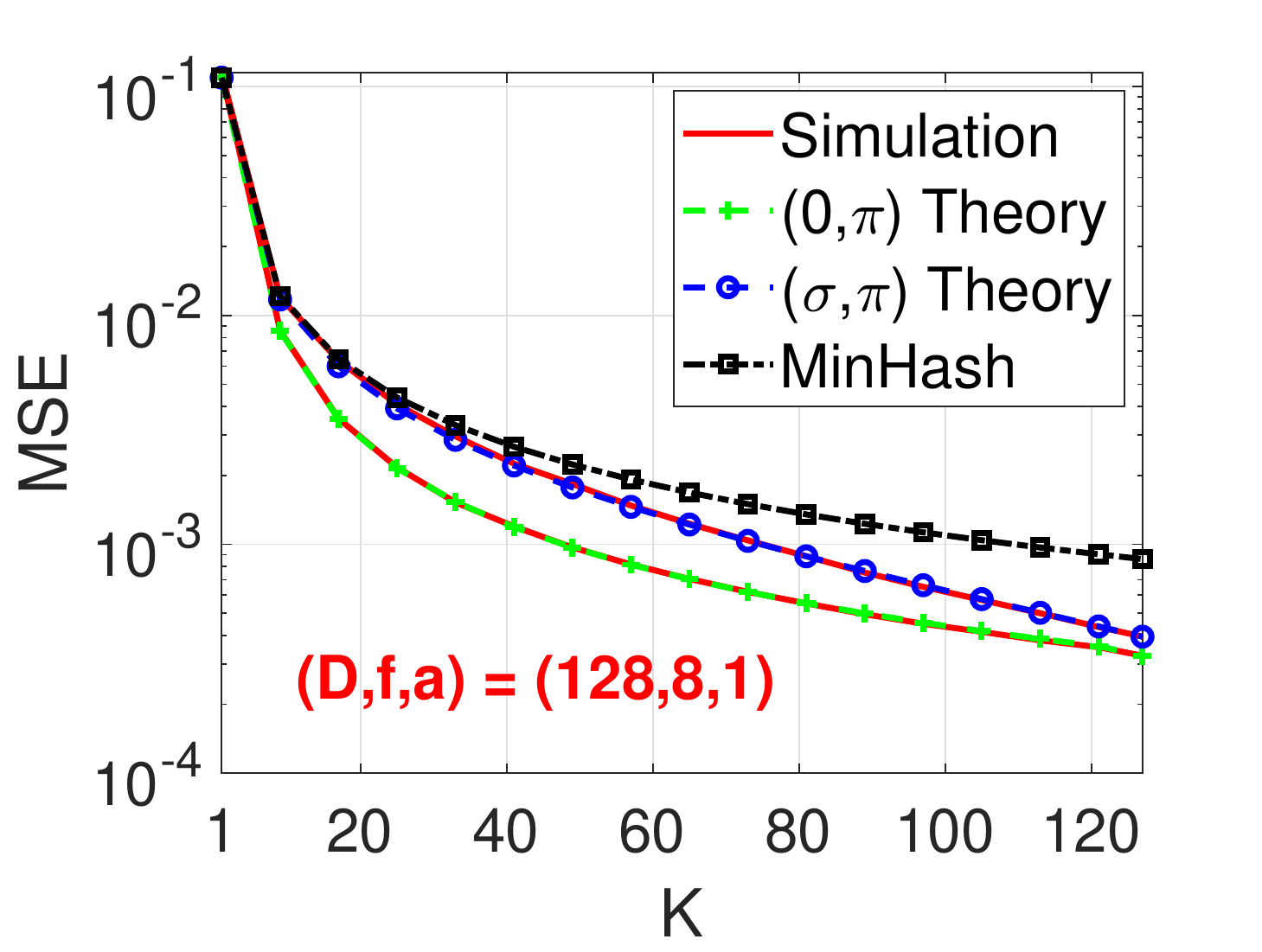}
		\includegraphics[width=2.7in]{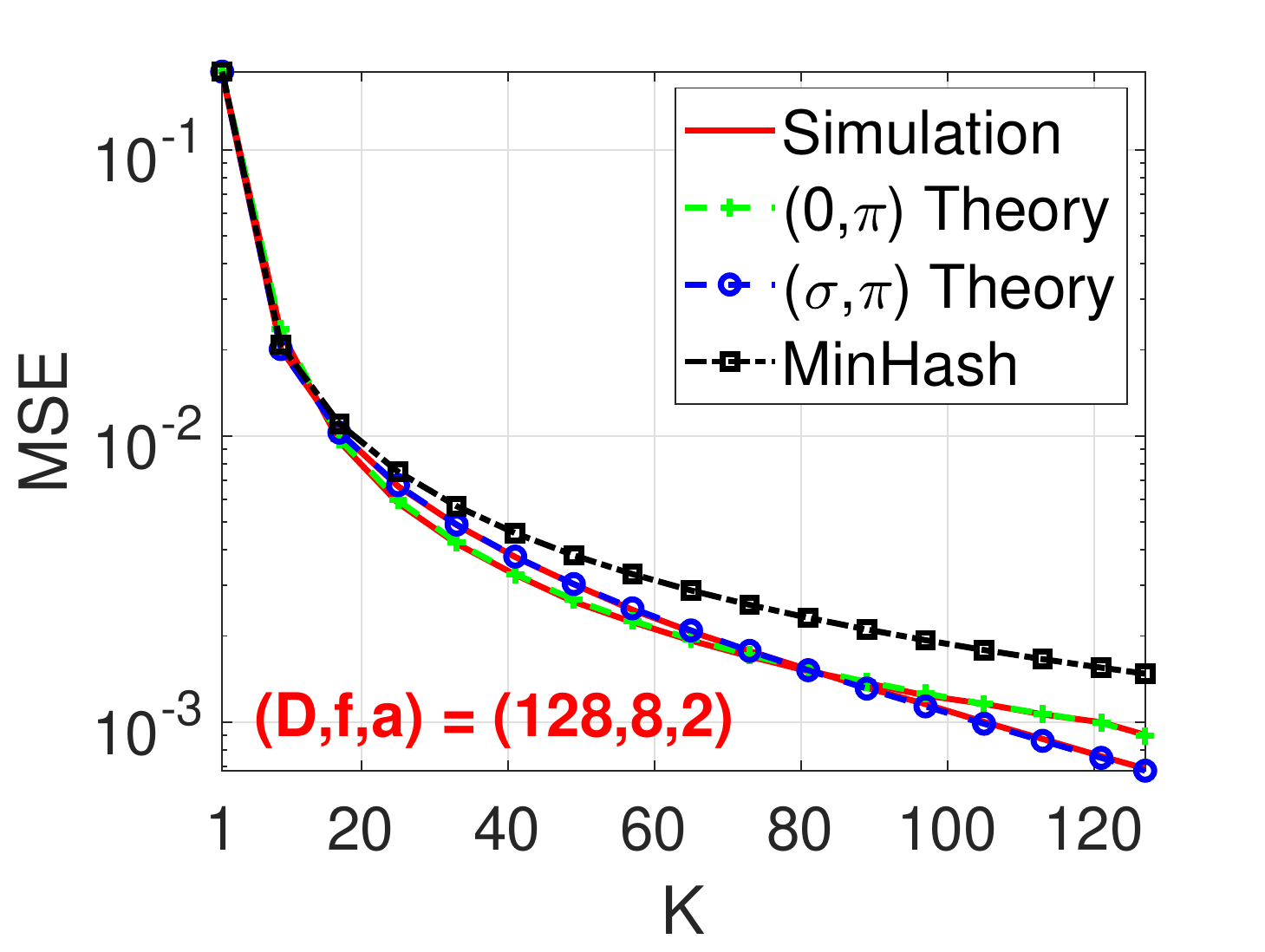}
		}
		\mbox{
		\includegraphics[width=2.7in]{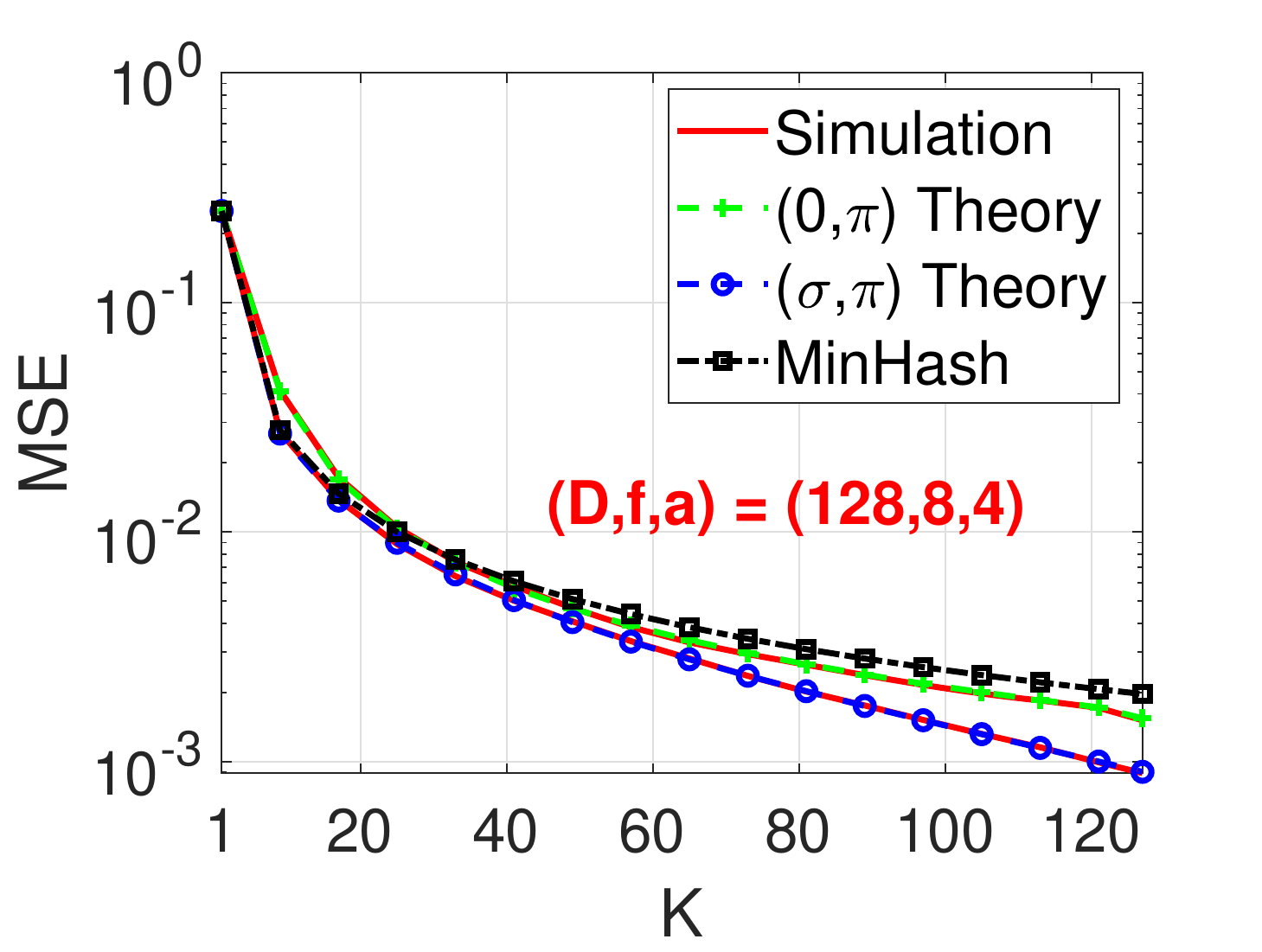}
		\includegraphics[width=2.7in]{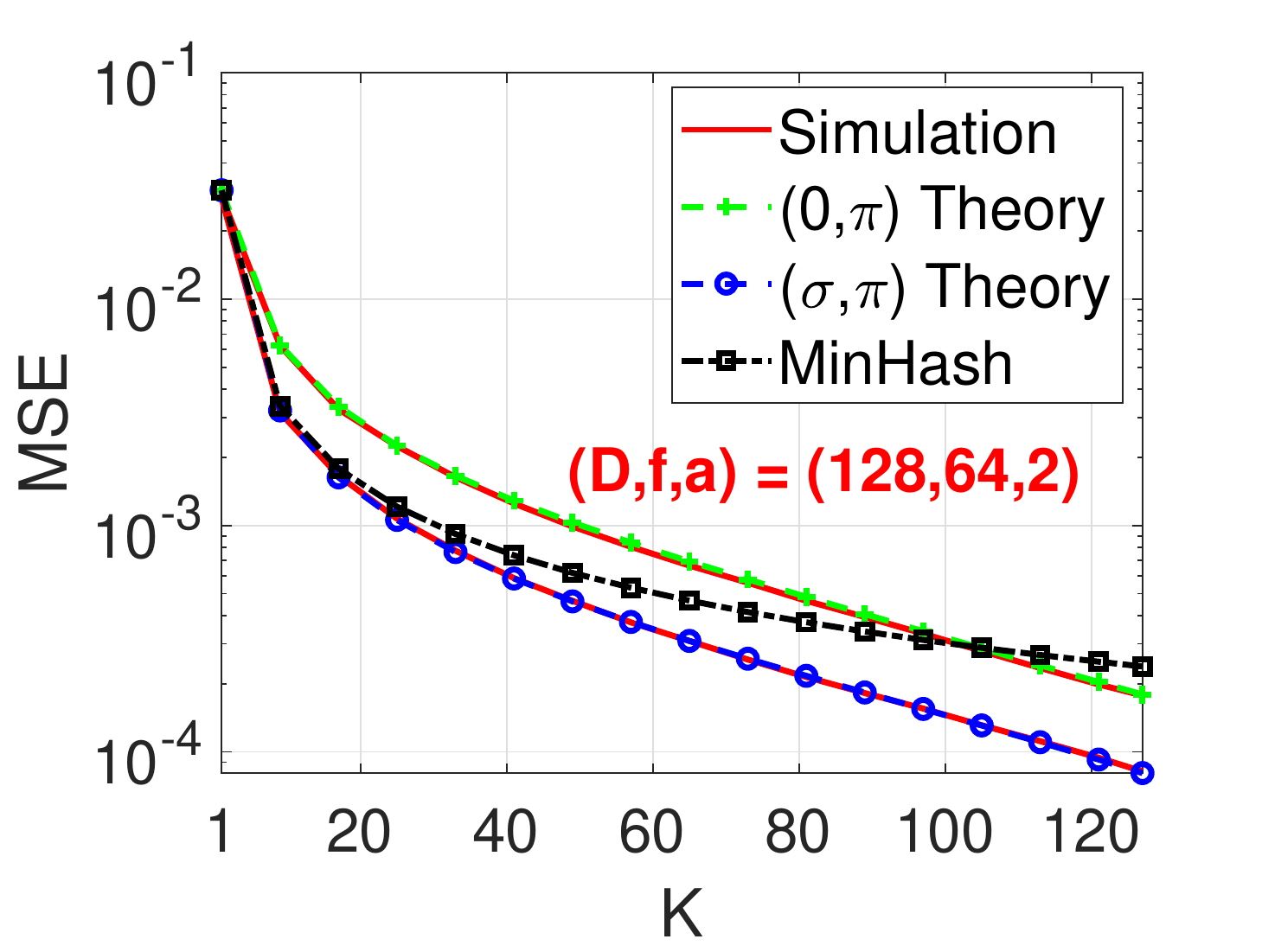}
		}
		\mbox{
		\includegraphics[width=2.7in]{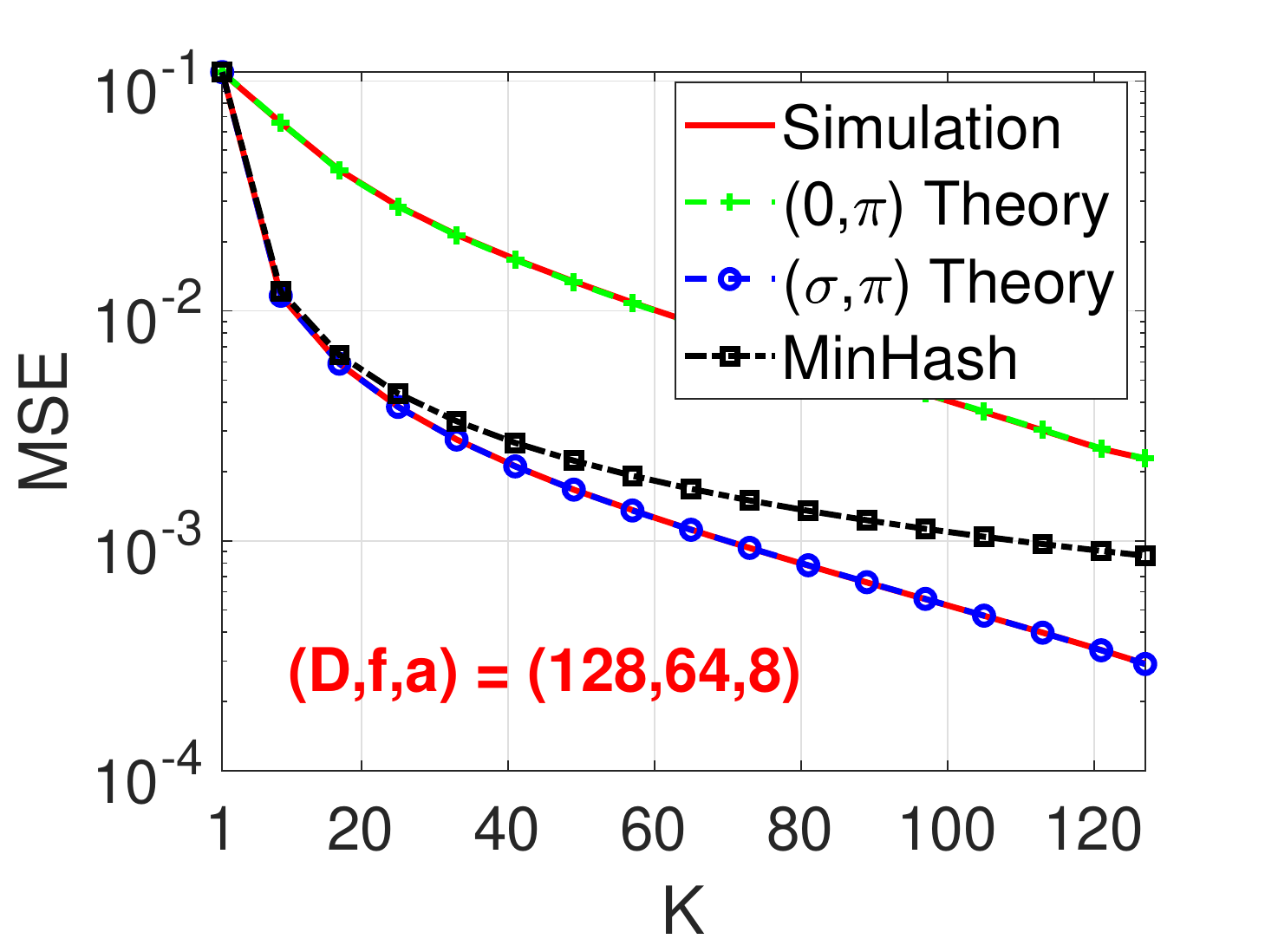}
		\includegraphics[width=2.7in]{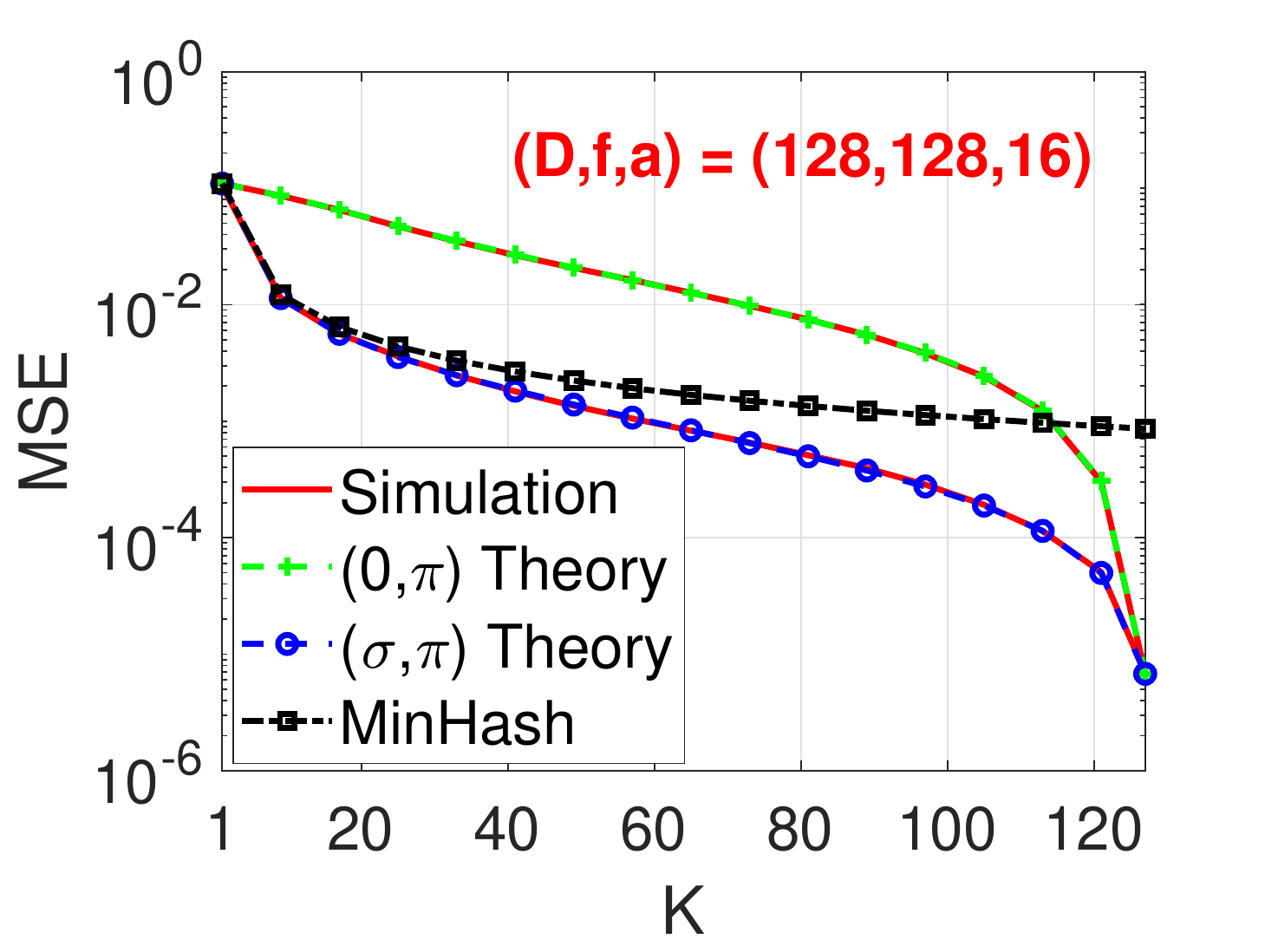}
		}
	\end{center}
	\vspace{-0.1in}
	\caption{Simulations for sanity check: Empirical vs. theoretical variance of $\hat J_{0,\pi}$ (C-MinHash-$(0,\pi)$) and $\hat J_{\sigma,\pi}$ (C-MinHash-$(\sigma,\pi)$), on synthetic binary data vector pairs with different data patterns.}
	\label{fig:sim-variance}
\end{figure}

A simulation study is conducted on synthetic data to verify the theoretical variances given by Theorem~\ref{theo:CMH-0,pi var} and Theorem~\ref{theo:CMH-sigma,pi var}. We simulate $D=128$ dimensional binary vector pairs $(\bm v,\bm w)$ with different combinations of $f$ and $a$. The vectors we generate has a special locational structure, where the location vector $\bm x$ is such that $a$ ``$O$'''s are followed by $(f-a)$ ``$\times$'''s and then followed by $(D-f)$ ``$-$'''s sequentially. We plot the empirical and theoretical mean square errors (MSE = variance + bias$^2$) in Figure~\ref{fig:sim-variance}:
\begin{itemize}
    \item The theoretical variance agrees with the empirical observation, as the curves overlap, confirming Theorem~\ref{theo:CMH-0,pi var} and Theorem~\ref{theo:CMH-sigma,pi var}. The variance reduction increases with larger $K$.

    \item $Var[\hat J_{\sigma,\pi}]$ (C-MinHash-$(\sigma,\pi)$) is always smaller than $Var[\hat J_{MH}]$, as stated by Theorem~\ref{theo:smaller-variance}. In contrast, $Var[\hat J_{0,\pi}]$ (C-MinHash-$(0,\pi)$) varies significantly depending on different data structures.
\end{itemize}

\newpage

\subsection{Jaccard Estimation on Text and Image Datasets} \label{sec:text-image data}

\begin{figure}[h]
	\begin{center}
\mbox{
		\includegraphics[width=2.7in]{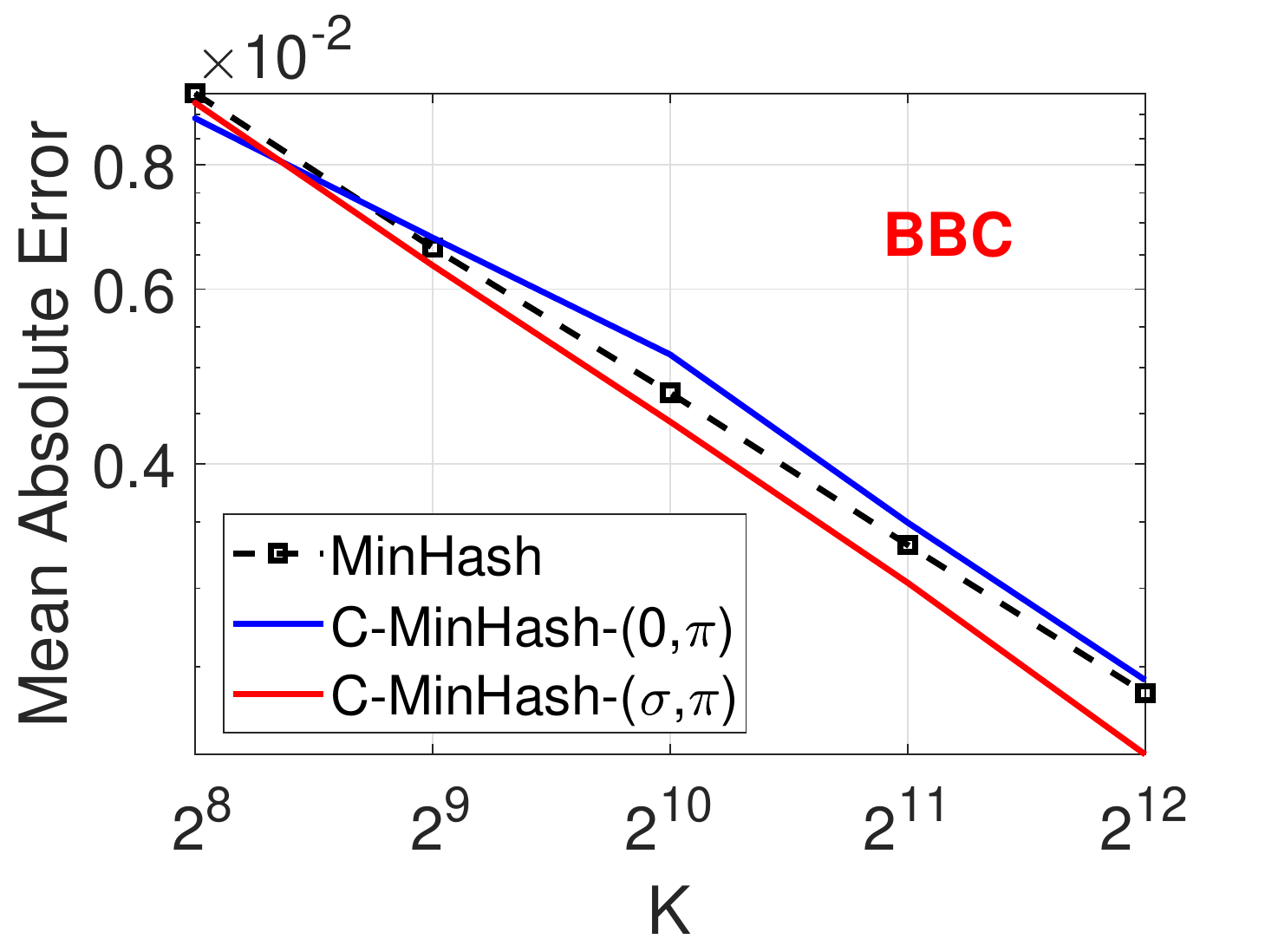}
		\includegraphics[width=2.7in]{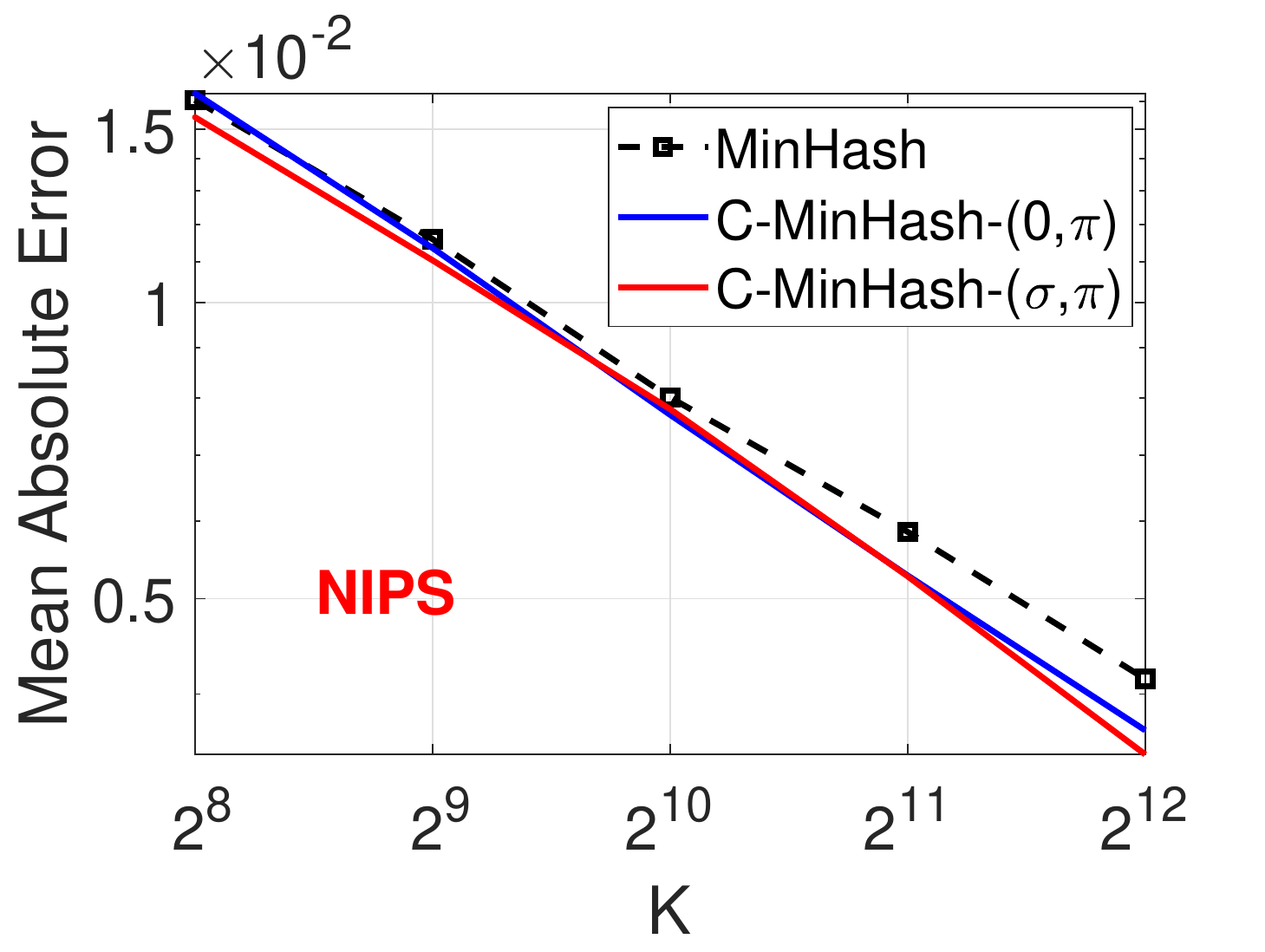}
	}
		\mbox{
		\includegraphics[width=2.7in]{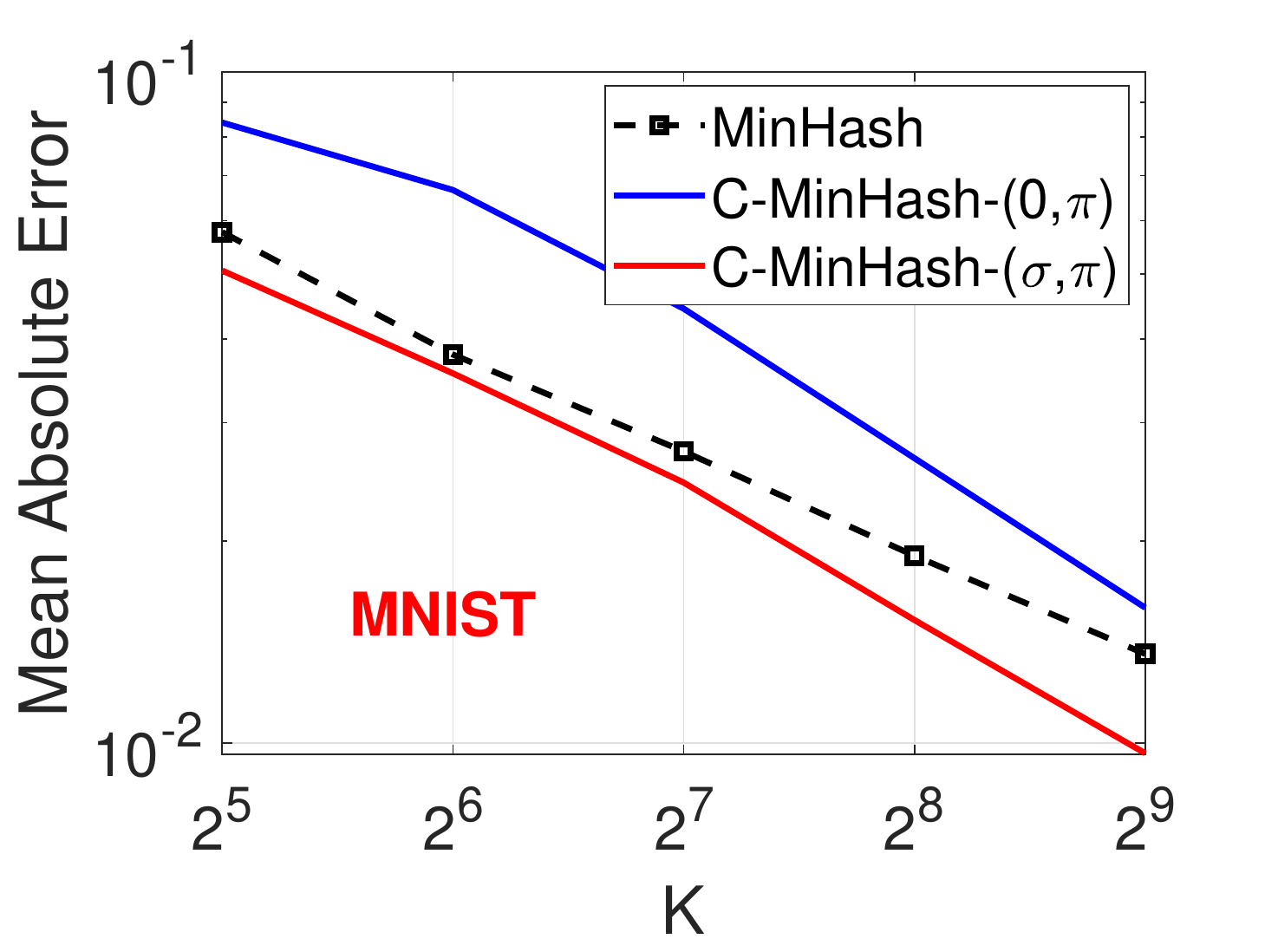}
		\includegraphics[width=2.7in]{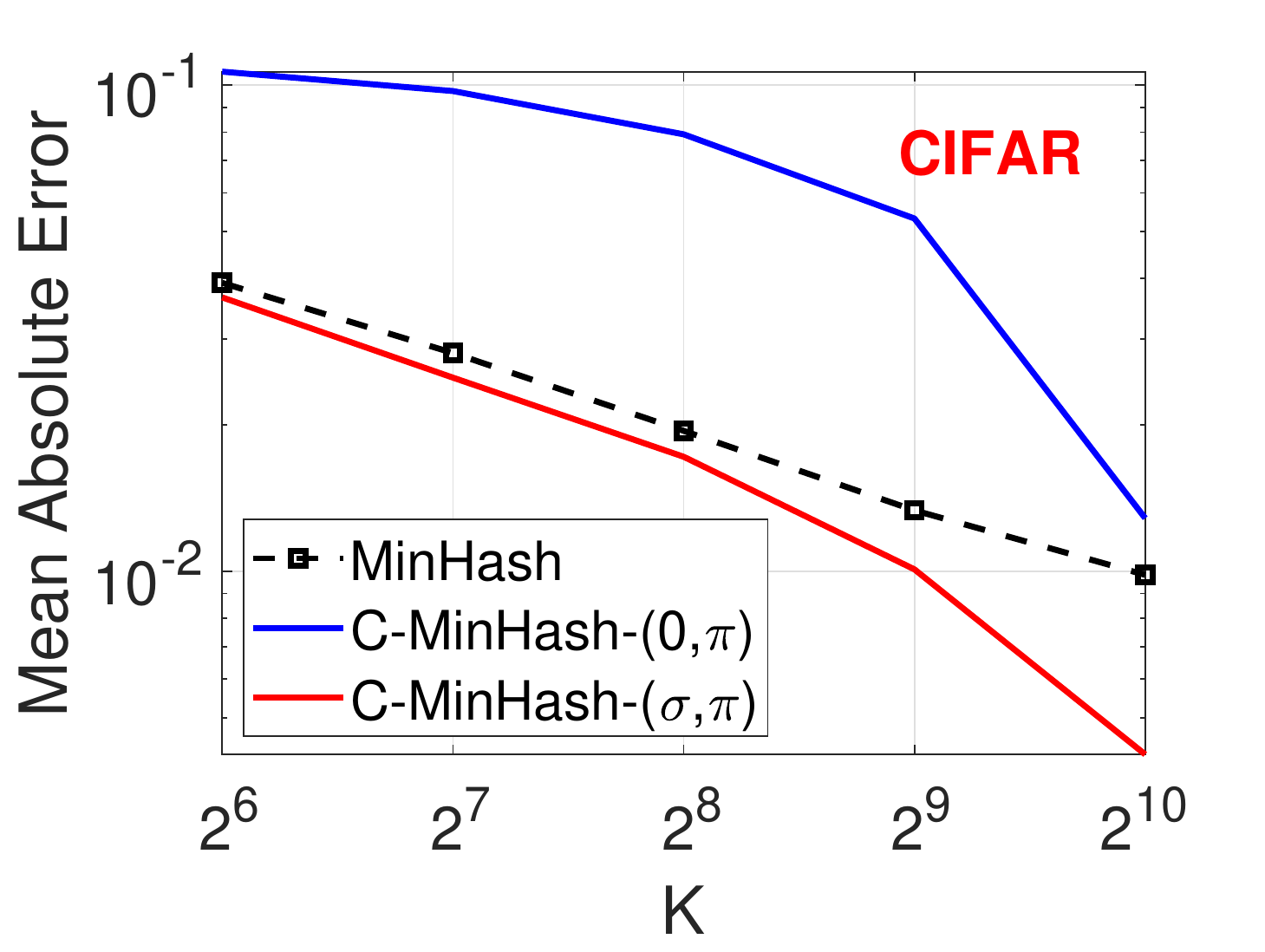}
		}
	\end{center}
	\vspace{-0.1in}
	\caption{Mean Absolute Error of MinHash and C-MinHash on real-world datasets.}
	\label{fig:error_experiment}
\end{figure}

We test C-MinHash on four commonly used datasets, including two text datasets: the NIPS full paper dataset from UCI repository~\citep{UCI}, and the BBC News dataset~\citep{greene06icml}, and two popular image datasets: the MNIST dataset~\citep{mnist} with hand-written digits, and  the CIFAR dataset~\citep{cifar} containing natural images. All the datasets are processed to be binary. For each dataset with $n$ data vectors, there are in total $n(n-1)/2$ data vector pairs. We estimate the Jaccard similarities for all the pairs and report the mean absolute errors (MAE). The results are averaged over 10 independent repetitions, for each dataset, as shown in Figure~\ref{fig:error_experiment}:

\begin{itemize}
    \item The MAE of C-MinHash-$(\sigma,\pi)$ is consistently smaller than that of MinHash, confirming our theoretical claim of improved estimation accuracy (Theorem~\ref{theo:smaller-variance}). We  also observe that the improvements become more substantial with larger $K$, which is consistent with the trend in Figure~\ref{fig:var-ratio}.

    \item
    Without the initial permutation $\sigma$, the accuracy of C-MinHash-$(0,\pi)$  is affected by the distribution of the original data, and it is worse than  C-MinHash-$(\sigma,\pi)$ on all these four datasets. One can also observe that  the performance of C-MinHash-$(0,\pi)$  on image data seems much worse than that on text data. We believe this is because the image datasets contain more structural patterns. This again confirms that the initial permutation $\sigma$ might be needed in practice.

\end{itemize}

\vspace{0.2in}

In summary, the simulation study has verified the correctness of our theoretical findings in Theorem~\ref{theo:CMH-0,pi var} and  Theorem~\ref{theo:CMH-sigma,pi var}. The experiments with Jaccard estimation on four datasets confirm that C-MinHash is more accurate than the original MinHash. The initial permutation $\sigma$ is recommended.

\section{Conclusion}

The method of {\em minwise hashing} (MinHash), from the seminal works of Broder and his colleagues, has become standard in industrial practice.  One  fundamental reason for its wide applicability is that the binary (0/1) high-dimensional representation is very convenient and  suitable for a wide range of practical scenarios. These days, (deep) learning for short representations has become popular, but it cannot replace the practice of using simple binary high-dimensional representations. It is  natural to view the world as a high-dimensional space and each object is mapped to this space as a sparse vector with only a small number of non-zero entries. While usually it is  challenging to assign a real-valued weight to each non-zero entry, engineering experience says that, with such a high-dimensional space, it is often the case that mainly the locations of the non-zero entries matter. We believe that the binary representations will still be widely used for a very long time.

\vspace{0.1in}

\noindent To estimate the Jaccard similarity $J$, if one hopes to achieve a strictly unbiased estimator $\hat J$ with the variance following exactly the binomial distribution, i.e., $\mathbb{E} [\hat J] = J$ and $Var[\hat J] = \frac{J(J-1)}{K}$, then one has to use $K$ independent permutations, where $K$, the number of hashes,   can be several hundreds or even thousands in practice. In this paper, we present the surprising theoretical results that, with merely 2 permutations, we still obtain an unbiased estimate of the Jaccard similarity with the variance strictly smaller than that of the original MinHash.  The initial permutation is applied to break whatever structure the original data may exhibit. The second permutation is re-used $K$ times in a circulant shifting fashion. Obviously the hash samples are no longer independent, but our analysis has shown that the estimation variance is actually smaller than that of the original MinHash, as confirmed by numerical experiments on simulated and real datasets.

\vspace{0.1in}

\noindent Practically speaking, our theoretical results presented in this paper may also reveal a useful direction for designing  hashing methods. For example, in many applications, using permutation vectors of length (e.g.,) $2^{30}$ might be sufficient. While it is perhaps unrealistic to store (e.g.,) $K=1024$ such permutation vectors in the memory, one can easily afford to store  two such permutations (even in the GPU memory). Using perfectly random permutations in lieu of approximate permutations would be able to simplify the design and analysis of randomized algorithms and ensure that the practical performance strictly matches the theory.

\newpage
\appendix
\renewcommand\thefigure{\thesection.\arabic{figure}}

\noindent\textbf{\LARGE Appendix}

\section{Proofs of Technical Results}
\vspace{0.1in}

\paragraph{Notations.}In our analysis, we will use $\mathbbm 1_s$ to denote $\mathbbm 1\{h_s(\bm v)=h_s(\bm w)\}$ in for $\forall 1\leq s\leq K$, where $h$ is the hash value. Given two data vectors $\bm v,\bm w\in \{0,1\}^D$. Recall in~\eqref{def:a, f}:
$a = \sum_{i=1}^D\mathbbm 1\{ \bm v_i = 1 \text{ and }  \bm w_i = 1\}, \
f = \sum_{i=1}^D\mathbbm 1\{ \bm v_i = 1 \text{ or }  \bm w_i = 1\}$.  Thus, the Jaccard similarity $J=a/f$. We also define $\tilde J=(a-1)/(f-1)$.

\vspace{0.1in}

\begin{definition}\label{append:def-1}
Let $\bm v,\bm w\in \{0,1\}^D$. Define the \textbf{location vector} as $\bm x\in\{O,\times,-\}^D$, with $\bm x_i$ being ``$O$'', ``$\times$'', ``$-$'' when $\bm v_i=\bm w_i=1$, $\bm v_i+\bm w_i=1$ and $\bm v_i=\bm w_i=0$, respectively.
\end{definition}

\vspace{0.1in}

\begin{definition}\label{append:def-sets}
For $A,B\in\{O,\times,-\}$, let $\{(i,j):(A,B)|\triangle\}$ denote a pair of indices $\{(i,j):(\bm x_i,\bm x_j)=(A,B),j-i=\triangle\}$. Define
\begin{align*}
    &\mathcal L_0(\triangle)=\{(i,j):(O,O)|\triangle\},\hspace{0.1in}\mathcal L_1(\triangle)=\{(i,j):(O,\times)|\triangle\},\hspace{0.1in}\mathcal L_2(\triangle)=\{(i,j):(O,-)|\triangle\},\\
    &\mathcal G_0(\triangle)=\{(i,j):(-,O)|\triangle\},\hspace{0.1in}\mathcal G_1(\triangle)=\{(i,j):(-,\times)|\triangle\}\hspace{0.1in}, \mathcal G_2(\triangle)=\{(i,j):(-,-)|\triangle\},\\
    &\mathcal H_0(\triangle)=\{(i,j):(\times ,O)|\triangle\},\hspace{0.1in}\mathcal H_1(\triangle)=\{(i,j): (\times ,\times)|\triangle\},\hspace{0.1in} \mathcal H_2(\triangle)=\{(i,j):(\times ,-)|\triangle\}.
\end{align*}
\end{definition}
\begin{remark}
For the ease of notation, by circulation we write $\bm x_j=\bm x_{j-D}$ when $D<j<2D$.
\end{remark}

One can easily verify that given fixed $a,f,D$, it holds that for $\forall 1\leq \triangle\leq K-1$,
\begin{align}
\begin{aligned}
    &|\mathcal L_0(\triangle)|+|\mathcal L_1(\triangle)|+|\mathcal L_2(\triangle)|=|\mathcal L_0(\triangle)|+|\mathcal G_0(\triangle)|+|\mathcal H_0(\triangle)|=a,\\
    &|\mathcal G_0(\triangle)|+|\mathcal G_1(\triangle)|+|\mathcal G_2(\triangle)|=|\mathcal L_2(\triangle)|+|\mathcal G_2(\triangle)|+|\mathcal H_2(\triangle)|=D-f,\\
    &|\mathcal H_0(\triangle)|+|\mathcal H_1(\triangle)|+|\mathcal H_2(\triangle)|=|\mathcal L_1(\triangle)|+|\mathcal G_1(\triangle)|+|\mathcal H_1(\triangle)|=f-a.
\end{aligned} \label{eqn:constraint append}
\end{align}
We will refer this as the intrinsic constraints on the size of above sets.

\subsection{Proof of Lemma \ref{lemma1}}  \label{sec:lemma1 proof}

\begin{manuallemma} {2.1}
 For any $1\leq s<t\leq K$ with $t-s=\triangle$, we have
\begin{equation*}
    \mathbb E_\pi\big[\mathbbm 1\{h_s(\bm v)=h_s(\bm w)\} \mathbbm 1\{h_t(\bm v)=h_t(\bm w)\}\big]=\frac{|\mathcal L_0(\triangle)|+(|\mathcal G_0(\triangle)|+|\mathcal L_2(\triangle)|)J}{f+|\mathcal G_0(\triangle)|+|\mathcal G_1(\triangle)|},
\end{equation*}
where the sets are defined in Definition~\ref{def-sets} and $h_s$, $h_t$ are the hash values as in Algorithm~\ref{alg:C-MinHash}.
\end{manuallemma}

\begin{proof}
To check whether a hash sample generated by MinHash collides (under some permutation $\pi$), it suffices to look at the permuted location vector $\bm x$. If a collision happens, after permuted by $\pi$, type ``$O$'' point must appear before the first ``$\times$'' point. That said, the minimal permutation index of ``$O$'' elements must be smaller than that of ``$\times$'' elements. If the hash sample does not collide, then the first ``$\times$'' must appear before the first ``$O$''. Note that ``$-$'' points does not affect the collision.

To compute the variance of the estimator, it suffices to compute $\mathbb E[\mathbbm 1_s\mathbbm 1_t]$. Let $\mathcal L$, $\mathcal G$ and $\mathcal H$ be the union of $\mathcal L$'s, $\mathcal G$'s and $\mathcal H$'s, respectively. In the following, we say that an index $i$ belongs to a set if $i$ is the first term of an element in that set. We have
\begin{equation*}
    |\mathcal L|=a,\quad |\mathcal H|=f-a,\quad |\mathcal G|=D-f.
\end{equation*}
One key observation is that, for a pair $(i,j)$ in above sets, the hash index $\pi_s(i)$ will be the hash index of $\pi_t(j)$. We begin by decomposing the expectation into
\begin{align}
    \mathbb E[\mathbbm 1_s\mathbbm 1_t]&=P[\text{collision}\ s, \text{collision}\ t] \nonumber\\
    &=\sum_{i_s^*\in \mathcal L}P[\text{collision}\ s\ \text{at}\ i_s^*, \text{collision}\ t] \nonumber\\
    &=\sum_{p=0}^2\sum_{i_s^*\in \mathcal L_p}P[\text{collision}\ s\ \text{at}\ i_s^*, \text{collision}\ t]. \label{eqn-1}
\end{align}
where $i_s^*$ is the location of the original ``$O$'' in vector $x$ that collides for $s$-th hash sample. It is different from the exact location of collision in $x(\pi_s)$. Note that the permutation is totally random, so the location of collision is independent of $\mathbbm 1_s$, and uniformly distributed among all type ``$O$'' pairs.\\

\textbf{1) When $i_{s}^*\in \mathcal L_0$.}
In this case, the minimum index of the type ``$O$'' pair in $x(\pi_s)$, $\pi_s(i_s^*)$, is shifted to another type ``$O$'' pair in $x(\pi_t)$. Therefore, the indices of pairs with the first element being ``$O$'' or ``$\times$'' originally in $x(\pi_s)$ will still be greater than $\pi_t(i_s^*)$. If sample $s$ collides at $i_s^*$, hash sample $t$ will collide when
\begin{enumerate}
    \item All the points in $\mathcal G_1$, after permutation $\pi_s$, is greater than $\pi_s(i_s^*)$. In this case, regardless of the permuted $\mathcal G_0$, hash $t$ will always collide.

    \item There exist points in $\mathcal G_1$ after permutation $\pi_s$ smaller than $\pi_s(i_s^*)$, and also points in $\mathcal G_0$ that is smaller than the minimum of permuted $\mathcal G_1$.
\end{enumerate}
Consequently, we have for $i_s^*\in\mathcal L_0$,
\begin{align}
    &P[\text{collision}\ s\ \text{at}\ i_s^*, \text{collision}\ t] \nonumber\\
    &=P[\pi_s(i_s^*)<\pi_s(i), \forall i\in \mathcal H\cup\mathcal L/i_s^*,\ \text{and}\ \min_{j\in\mathcal G_1}\pi_s(j) >\pi_s(i_s^*)] \nonumber\\
    &\hspace{0.6in}+P[\pi_s(i_s^*)<\pi_s(i),\forall i\in \mathcal H\cup\mathcal L/i_s^*,\ \text{and}\ \min_{j\in\mathcal G_0}\pi_s(j)<\min_{j\in\mathcal G_1}\pi_s(j)<\pi_s(i_s^*)] \nonumber\\
    &=\frac{1}{a}\cdot \frac{a}{f+|\mathcal G_1|}+\frac{|\mathcal G_0|}{f+|\mathcal G_0|+|\mathcal G_1|}\cdot \frac{|\mathcal G_1|}{f+|\mathcal G_1|} \cdot\frac{a}{f}\cdot \frac{1}{a} \nonumber\\
    &=\frac{1}{f+|\mathcal G_1|}+\frac{|\mathcal G_0|\cdot |\mathcal G_1|}{(f+|\mathcal G_0|+|\mathcal G_1|)(f+|\mathcal G_1|)f}. \label{eqn-case1}
\end{align}
This probability holds for $\forall i_s^*\in\mathcal L_0$.\\

\textbf{2) When $i_{s}^*\in \mathcal L_1$.}
Similarly, we consider the condition where $i_s^*\in\mathcal L_1$, and both hash samples collide. In this case, $\pi_s(i_s^*)$ would be shifted to a ``$\times$'' pair in $x(\pi_t)$. That is, the indices of pairs with the first element being ``$O$'' or ``$\times$'' originally in $x(\pi_s)$ will all become greater than $\pi_s(i_s^*)$, which now is the location of a ``$\times$'' pair in $x(\pi_t)$. Thus, to make hash $t$ collide, we will need:
\begin{itemize}
   \item At least one point from $\mathcal G_0$ is smaller than any other points in $\mathcal H\cup\mathcal L\cup \mathcal G_1$ after permutation $\pi_s$.
\end{itemize}
Therefore, for any $i_s^*\in\mathcal L_1$,
\begin{align}
    &P[\text{collision}\ s\ \text{at}\ i_s^*, \text{collision}\ t] \nonumber\\
    &=P[\pi_s(i_s^*)<\pi_s(i), \forall i\in \mathcal H\cup\mathcal L/i_s^*,\ \text{and}\ \min_{j\in\mathcal G_0}\pi_s(j)<\min\{\pi_s(i_s^*),\min_{j\in\mathcal G_1}\pi_s(j)\}] \nonumber\\
    &=\frac{|\mathcal G_0|}{f+|\mathcal G_0|+|\mathcal G_1|}\cdot \frac{a}{f}\cdot \frac{1}{a} \nonumber\\
    &=\frac{|\mathcal G_0|}{(f+|\mathcal G_0|+|\mathcal G_1|)f}, \label{eqn-case2}
\end{align}
which is true for $\forall i_s^*\in\mathcal L_1$.\\

\textbf{3) When $i_{s}^*\in \mathcal L_2$.}

In this scenario, $\pi_s(i_s^*)$ would be shifted to a ``$-$'' pair in $x(\pi_t)$. Therefore, if hash $s$ collides, hash $t$ will also collide when:
\begin{itemize}
   \item After applying $\pi_s$, the minimum of $\mathcal L_0\cup \mathcal H_0\cup \mathcal G_0$ is smaller than the minimum of $\mathcal L_1\cup \mathcal H_1\cup \mathcal G_1$.
\end{itemize}
Thus, we obtain that for any $i_s^*\in \mathcal L_2$,
\begin{align*}
    &P[\text{collision}\ s\ \text{at}\ i_s^*, \text{collision}\ t]\\
    &=P[\pi_s(i_s^*)<\pi_s(i), \forall i\in \mathcal H\cup\mathcal L/i_s^*,\ \text{and}\ \min_{j\in\mathcal L_0\cup \mathcal G_0\cup \mathcal H_0}\pi_s(j)<\min_{j\in\mathcal L_1\cup \mathcal G_1\cup \mathcal H_1}\pi_s(j)]\\
    &\triangleq P[\Omega].
\end{align*}
Let $\mathbbm 1_{s,i_s^*}$ denote the event $\{\pi_s(i_s^*)<\pi_s(i), \forall i\in \mathcal H\cup\mathcal L/i_s^*\}$. Then $\Omega$ can be separated into the following several cases:
\begin{enumerate}
    \item $\Omega_1$: $\mathbbm 1_{s,i_s^*}$, $\min_{j\in\mathcal L_0\cup\mathcal H_0}\pi_s(j)<\min_{j\in\mathcal L_1\cup\mathcal H_1}\pi_s(j)$, and $\min_{j\in\mathcal L_0\cup\mathcal H_0}\pi_s(j)<\min_{j\in\mathcal G_1}\pi_s(j)$.

    \item $\Omega_2$: $\mathbbm 1_{s,i_s^*}$, $\min_{j\in\mathcal L_0\cup\mathcal H_0}\pi_s(j)<\min_{j\in\mathcal L_1\cup\mathcal H_1}\pi_s(j)$, and $\min_{j\in\mathcal L_0\cup\mathcal H_0}\pi_s(j)>\min_{j\in\mathcal G_1}\pi_s(j)>\min_{j\in\mathcal G_0}\pi_s(j)>\pi_s(i_s^*)$.

    \item $\Omega_3$: $\mathbbm 1_{s,i_s^*}$, $\min_{j\in\mathcal L_0\cup\mathcal H_0}\pi_s(j)<\min_{j\in\mathcal L_1\cup\mathcal H_1}\pi_s(j)$, and $\min_{j\in\mathcal L_0\cup\mathcal H_0}\pi_s(j)>\min_{j\in\mathcal G_1}\pi_s(j)>\pi_s(i_s^*)>\min_{j\in\mathcal G_0}\pi_s(j)$.

    \item $\Omega_4$: $\mathbbm 1_{s,i_s^*}$, $\min_{j\in\mathcal L_0\cup\mathcal H_0}\pi_s(j)<\min_{j\in\mathcal L_1\cup\mathcal H_1}\pi_s(j)$, and $\min_{j\in\mathcal L_0\cup\mathcal H_0}\pi_s(j)>\pi_s(i_s^*)>\min_{j\in\mathcal G_1}\pi_s(j)>\min_{j\in\mathcal G_0}\pi_s(j)$.

    \item $\Omega_5$: $\mathbbm 1_{s,i_s^*}$, $\min_{j\in\mathcal L_0\cup\mathcal H_0}\pi_s(j)>\min_{j\in\mathcal L_1\cup\mathcal H_1}\pi_s(j)$, and
    $\pi_s(i_s^*)<\min_{j\in\mathcal G_0}\pi_s(j)<\min_{j\in\mathcal L_1\cup\mathcal H_1\cup \mathcal G_1}\pi_s(j)$.

    \item $\Omega_6$: $\mathbbm 1_{s,i_s^*}$, $\min_{j\in\mathcal L_0\cup\mathcal H_0}\pi_s(j)>\min_{j\in\mathcal L_1\cup\mathcal H_1}\pi_s(j)$, and
    $\min_{j\in\mathcal G_0}\pi_s(j)<\pi_s(i_s^*)<\min_{j\in\mathcal L_1\cup\mathcal H_1\cup \mathcal G_1}\pi_s(j)$.
\end{enumerate}
We can compute the probability of each event as
\begin{align*}
    P[\Omega_1]&=\frac{1}{a}\cdot \frac{a}{f+|\mathcal G_1|}\cdot \frac{|\mathcal L_0|+|\mathcal H_0|}{|\mathcal L_0|+|\mathcal H_0|+|\mathcal L_1|+|\mathcal H_1|+|\mathcal G_1|},\\
    &=\frac{a-|\mathcal G_0|}{(f-|\mathcal G_0|)(f+|\mathcal G_1|)},\\
    P[\Omega_2]&=\frac{1}{a}\cdot \frac{a}{f+|\mathcal G_0|+|\mathcal G_1|}\cdot \frac{|\mathcal G_0|}{|\mathcal G_0|+|\mathcal G_1|+|\mathcal L_0|+|\mathcal H_0|+|\mathcal L_1|+|\mathcal H_1|}\\
    &\hspace{0.8in} \cdot \frac{|\mathcal G_1|}{|\mathcal L_0|+|\mathcal H_0|+|\mathcal L_1|+|\mathcal H_1|+|\mathcal G_1|}\cdot \frac{|\mathcal L_0|+|\mathcal H_0|}{|\mathcal L_0|+|\mathcal H_0|+|\mathcal L_1|+|\mathcal H_1|}\\
    &=\frac{1}{f+|\mathcal G_0|+|\mathcal G_1|}\cdot \frac{|\mathcal G_0|}{f}\cdot \frac{|\mathcal G_1|}{f-|\mathcal G_0|}\cdot \frac{a-|\mathcal G_0|}{f-|\mathcal G_0|-|\mathcal G_1|}\\
    &=\frac{|\mathcal G_0|\cdot |\mathcal G_1|\cdot (a-|\mathcal G_0|)}{(f+|\mathcal G_0|+|\mathcal G_1|)(f-|\mathcal G_0|)(f-|\mathcal G_0|-|\mathcal G_1|)f},\\
    P[\Omega_3]&=\frac{|\mathcal G_0|}{f+|\mathcal G_0|+|\mathcal G_1|}\cdot \frac{1}{f+|\mathcal G_1|}\cdot \frac{|\mathcal G_1|}{|\mathcal L_0|+|\mathcal H_0|+|\mathcal L_1|+|\mathcal H_1|+|\mathcal G_1|}\cdot \frac{|\mathcal L_0|+|\mathcal H_0|}{|\mathcal L_0|+|\mathcal H_0|+|\mathcal L_1|+|\mathcal H_1|}\\
    &=\frac{|\mathcal G_0|\cdot |\mathcal G_1|\cdot (a-|\mathcal G_0|)}{(f+|\mathcal G_0|+|\mathcal G_1|)(f+|\mathcal G_1|)(f-|\mathcal G_0|)(f-|\mathcal G_0|-|\mathcal G_1|)},\\
    P[\Omega_4]&=\frac{|\mathcal G_0|}{f+|\mathcal G_0|+|\mathcal G_1|}\cdot \frac{|\mathcal G_1|}{f+|\mathcal G_1|}\cdot \frac{1}{f}\cdot \frac{|\mathcal L_0|+|\mathcal H_0|}{|\mathcal L_0|+|\mathcal H_0|+|\mathcal L_1|+|\mathcal H_1|}\\
    &=\frac{|\mathcal G_0|\cdot |\mathcal G_1|\cdot (a-|\mathcal G_0|)}{(f+|\mathcal G_0|+|\mathcal G_1|)(f+|\mathcal G_1|)(f-|\mathcal G_0|-|\mathcal G_1|)f},\\
    P[\Omega_5]&=\frac{1}{f+|\mathcal G_0|+|\mathcal G_1|}\cdot \frac{|\mathcal G_0|}{|\mathcal G_0|+|\mathcal G_1|+|\mathcal L_0|+|\mathcal H_0|+|\mathcal L_1|+|\mathcal H_1|}\cdot \frac{|\mathcal L_1|+|\mathcal H_1|}{|\mathcal L_0|+|\mathcal H_0|+|\mathcal L_1|+|\mathcal H_1|}\\
    &=\frac{|\mathcal G_0|\cdot (f-a-|\mathcal G_1|)}{(f+|\mathcal G_0|+|\mathcal G_1|)(f-|\mathcal G_0|-|\mathcal G_1|)f},\\
    P[\Omega_6]&=\frac{|\mathcal G_0|}{f+|\mathcal G_0|+|\mathcal G_1|}\cdot \frac{1}{f}\cdot \frac{|\mathcal L_1|+|\mathcal H_1|}{|\mathcal L_0|+|\mathcal H_0|+|\mathcal L_1|+|\mathcal H_1|}\\
    &=\frac{|\mathcal G_0|\cdot (f-a-|\mathcal G_1|)}{(f+|\mathcal G_0|+|\mathcal G_1|)(f-|\mathcal G_0|-|\mathcal G_1|)f}.
\end{align*}
Note that
\begin{align*}
    &P[\Omega_2]+P[\Omega_3]+P[\Omega_4]\\
    &=\frac{|\mathcal G_0|\cdot |\mathcal G_1|\cdot (a-|\mathcal G_0|)}{(f+|\mathcal G_0|+|\mathcal G_1|)(f-|\mathcal G_0|-|\mathcal G_1|)}\left[\frac{1}{(f-|\mathcal G_0|)f}+\frac{1}{(f-|\mathcal G_0|)(f+|\mathcal G_1|)}+\frac{1}{f(f+|\mathcal G_1|)}   \right]\\
    &=\frac{|\mathcal G_0|\cdot |\mathcal G_1|\cdot (a-|\mathcal G_0|)(3f-|\mathcal G_0|+|\mathcal G_1|)}{(f+|\mathcal G_0|+|\mathcal G_1|)(f-|\mathcal G_0|-|\mathcal G_1|)f(f-|\mathcal G_0|)(f+|\mathcal G_1|)}.
\end{align*}
Summing up all the terms together, we obtain $P[\Omega]$ as
\begin{align}
    \sum_{n=1}^6 P[\Omega_n]&=\frac{f(f+|\mathcal G_0|+|\mathcal G_1|)(f-|\mathcal G_0|-|\mathcal G_1|)(a-|\mathcal G_0|)+|\mathcal G_0||\mathcal G_1|(a-|\mathcal G_0|)(3f-|\mathcal G_0|+|\mathcal G_1|)}{(f+|\mathcal G_0|+|\mathcal G_1|)(f-|\mathcal G_0|-|\mathcal G_1|)(f-|\mathcal G_0|)(f+|\mathcal G_1|)f} \nonumber\\
    &\hspace{0.6in}  +\frac{2|\mathcal G_0|(f-a-|\mathcal G_1|)(f-|\mathcal G_0|)(f+|\mathcal G_1|)}{(f+|\mathcal G_0|+|\mathcal G_1|)(f-|\mathcal G_0|-|\mathcal G_1|)(f-|\mathcal G_0|)(f+|\mathcal G_1|)f} \nonumber\\
    &=\frac{(a-|\mathcal G_0|)(f+|\mathcal G_0|-|\mathcal G_1|)(f-|\mathcal G_0|)(f+|\mathcal G_1|)+2|\mathcal G_0|(f-a-|\mathcal G_1|)(f-|\mathcal G_0|)(f+|\mathcal G_1|)}{(f+|\mathcal G_0|+|\mathcal G_1|)(f-|\mathcal G_0|-|\mathcal G_1|)(f-|\mathcal G_0|)(f+|\mathcal G_1|)f} \nonumber\\
    &=\frac{(a+|\mathcal G_0|)(f-|\mathcal G_0|-|\mathcal G_1|)(f-|\mathcal G_0|)(f+|\mathcal G_1|)}{(f+|\mathcal G_0|+|\mathcal G_1|)(f-|\mathcal G_0|-|\mathcal G_1|)(f-|\mathcal G_0|)(f+|\mathcal G_1|)f} \nonumber\\
    &=\frac{a+|\mathcal G_0|}{(f+|\mathcal G_0|+|\mathcal G_1|)f}, \label{eqn-case3}
\end{align}
which holds for $\forall i_s^*\in\mathcal L_2$. Now combining (\ref{eqn-case1}), (\ref{eqn-case2}), (\ref{eqn-case3}) with (\ref{eqn-1}), we obtain
\begin{align}
    \mathbb E[\mathbbm 1_s\mathbbm 1_t]&=\frac{|\mathcal L_0|}{f+|\mathcal G_1|}+\frac{|\mathcal G_0||\mathcal G_1||\mathcal L_0|}{(f+|\mathcal G_0|+|\mathcal G_1|)(f+|\mathcal G_1|)f}+\frac{|\mathcal G_0||\mathcal L_1|}{(f+|\mathcal G_0|+|\mathcal G_1|)f}+\frac{(a+|\mathcal G_0|)|\mathcal L_2|}{(f+|\mathcal G_0|+|\mathcal G_1|)f}. \label{eqn:E_st}
\end{align}
Here, recall that the sets are associated with all $1\leq s<t\leq K$ such that $\triangle=t-s$. Using the intrinsic constraints (\ref{eqn:constraint append}), after some calculation we can simplify (\ref{eqn:E_st}) as
\begin{equation*}
    \mathbb E_\pi[\mathbbm 1_s\mathbbm 1_t]=\frac{|\mathcal L_0|}{f+|\mathcal G_0|+|\mathcal G_1|}+\frac{a(|\mathcal G_0|+|\mathcal L_2|)}{(f+|\mathcal G_0|+|\mathcal G_1|)f},
\end{equation*}
which completes the proof.
\end{proof}

\subsection{Proof of Theorem \ref{theo:CMH-0,pi var}} \label{sec:theo:CMH-0,pi var proof}

\begin{manualtheorem}{2.2}
Under the same setting as in Lemma~\ref{lemma1}, the variance of $\hat J_{0,\pi}$~is
\begin{align*}
    Var[\hat J_{0,\pi}]=\frac{J}{K}+\frac{2\sum_{s=2}^{K}(s-1)\Theta_{K-s+1}}{K^2}-J^2,
\end{align*}
where $\Theta_\triangle \triangleq E_\pi\big[\mathbbm 1\{h_s(\bm v)=h_s(\bm w)\} \mathbbm 1\{h_t(\bm v)=h_t(\bm w)\}\big]$ as in Lemma~\ref{lemma1} with any $t-s=\triangle$.
\end{manualtheorem}

\begin{proof}
By the expansion of variance formula, since $\mathbb E[\mathbbm 1_s^2]=\mathbb E[\mathbbm 1_s]=J$, we have
\begin{equation}
    Var[\hat J_{0,\pi}]=\frac{J}{K}+\frac{\sum_{s=1}^K\sum_{t\neq s}^K\mathbb E[\mathbbm 1_s\mathbbm 1_t]}{K^2}-J^2. \label{eqn-2}
\end{equation}
Note here that for $\forall t>s$, the $t$-th hash sample uses $\pi_t$ as the permutation, which is shifted rightwards by $\triangle=t-s$ from $\pi_s$. Thus, we have $\mathbb E[\mathbbm 1_s\mathbbm 1_t]=\mathbb E[\mathbbm 1_{s-i}\mathbbm 1_{t-i}]$ for $\forall 0<i<s\wedge t$, which implies $\mathbb E[\mathbbm 1_s\mathbbm 1_t]=\mathbb E[\mathbbm 1_1\mathbbm 1_{t-s+1}]$, $\forall s<t$. Since by assumption $K\leq D$, we have
\begin{align}
    \sum_{s}^K\sum_{t\neq s}^K\mathbb E[\mathbbm 1_s\mathbbm 1_t]&=2\mathbbm E\big[(\mathbbm 1_1\mathbbm 1_2+\mathbbm 1_1\mathbbm 1_3+...+\mathbbm 1_1\mathbbm 1_K)+(\mathbbm 1_2\mathbbm 1_3+...+\mathbbm 1_2\mathbbm 1_K)+...+\mathbbm 1_{K-1}\mathbbm 1_K\big] \nonumber\\
    &=2\mathbb E\big[(\mathbbm 1_1\mathbbm 1_2+\mathbbm 1_1\mathbbm 1_3+...+\mathbbm 1_1\mathbbm 1_K)+(\mathbbm 1_1\mathbbm 1_2+...+\mathbbm 1_1\mathbbm 1_{K-1})+...+\mathbbm 1_1\mathbbm 1_2\big] \nonumber\\
    &=2\sum_{s=2}^{K}(s-1)\mathbb E[\mathbbm 1_1\mathbbm 1_{K-s+2}] \nonumber\\
    &\triangleq 2\sum_{s=2}^{K}(s-1)\Theta_{K-s+1}. \label{eqn-3}
\end{align}
Finally, integrating (\ref{eqn-2}),  (\ref{eqn-3}) and Lemma~\ref{lemma1} completes the proof.
\end{proof}

\subsection{Proof of Theorem \ref{theo:CMH-sigma,pi var}}  \label{sec:theo:CMH-sigma,pi var proof}

\begin{manualtheorem}{3.1}
Let $a,f$ be defined as in (\ref{def:a, f}). When $0<a<f\leq D$ ($J\notin \{0,1\}$), we have
\begin{align}
    Var[\hat J_{\sigma,\pi}]=\frac{J}{K}+\frac{(K-1)\tilde{\mathcal E}}{K}-J^2,
\end{align}
where with $l=\max(0,D-2f+a)$, and
\begin{align}\notag
    \tilde{\mathcal E}=\sum_{\{l_0,l_2,g_0,g_1\}}\Bigg\{&\left(\frac{l_0}{f+g_0+g_1}+\frac{a(g_0+l_2)}{(f+g_0+g_1)f}\right) \\\label{eqn:E app} &\times\sum_{s=l}^{D-f-1}\frac{\binom{D-f}{s}}{\binom{D-a-1}{D-f-1}} \frac{\binom{f-a-1}{D-f-s-1}\binom{s}{n_1}\binom{D-f-s}{n_2}\binom{D-f-s}{n_3}\binom{f-a-(D-f-s)}{n_4}\binom{a-1}{a-l_1-l_2}}{\binom{D-1}{a}}\Bigg\}.
\end{align}
The feasible set $\{l_0,l_2,g_0,g_1\}$ satisfies the intrinsic constraints (\ref{eqn:constraint}), and
\begin{align*}
    &n_1=g_0-(D-f-s-g_1), \hspace{0.33in} n_2=D-f-s-g_1,\\
    &n_3=l_2-g_0+(D-f-s-g_1), \hspace{0.08in} n_4=l_1-(D-f-s-g_1).
\end{align*}
Also, when $a=0$ or $f=a$ ($J=0$ or $J=1$), we have $Var[\hat J_{\sigma,\pi}]=0$.
\end{manualtheorem}

\begin{proof}
Similar to the proof of Theorem~\ref{theo:CMH-0,pi var}, we denote $\Theta_\triangle=\mathbb E_{\sigma,\pi}[\mathbbm 1_s\mathbbm 1_t]$ with $|t-s|=\triangle$. Note that now the expectation is taken w.r.t. both two independent permutations $\sigma$ and $\pi$. Since $\sigma$ is random, we know that $\Theta_1=\Theta_2=\dots=\Theta_{K-1}$. Then by the variance formula, we have
\begin{align} \label{eqn:var-formula}
    Var[\hat J_{\sigma,\pi}]=\frac{J^2}{K}-\frac{(K-1)\Theta_1}{K}-J^2
\end{align}
Hence, it suffices to consider $\Theta_1$. In this proof, we will set $\triangle=1$ and drop the notation $\triangle$ for conciseness, and denote $\tilde{\mathcal E}=\Theta_1$ from now on. First, we note that Lemma~\ref{lemma1} gives the desired quantity conditional on $\sigma$. By the law of total probability, we have
\begin{align}
    \tilde{\mathcal E}=\mathbb E_{\sigma}\left[\frac{|\mathcal L_0|}{f+|\mathcal G_0|+|\mathcal G_1|}+\frac{a(|\mathcal G_0|+|\mathcal L_2|)}{(f+|\mathcal G_0|+|\mathcal G_1|)f}\right],  \label{eqn:condi-var}
\end{align}
where the sizes of sets are random depending on the initial permutation $\sigma$ (i.e. counted after permuting by~$\sigma$). As a result, the problem turns into deriving the distribution of $|\mathcal L_0|,|\mathcal L_1|,|\mathcal L_2|,|\mathcal G_0|$ and $|\mathcal G_1|$ under random permutation $\sigma$, and then taking expectation of (\ref{eqn:condi-var}) with respect to this additional randomness.\\

When $a=0$, we know that $|\mathcal L_0|=|\mathcal L_2|=|\mathcal G_0|=0$, hence the expectation $\tilde{\mathcal E}$ is trivially 0. Thus, the $Var[\hat J_{\sigma,\pi}]=0$. When $f=a$, $|\mathcal G_1|=0$, and the constraint on the sets becomes
\begin{align*}
    &|\mathcal L_0|+|\mathcal G_0|=|\mathcal L_0|+|\mathcal L_2|=f,\\
    &|\mathcal L_2|+|\mathcal G_2|=|\mathcal G_0|+|\mathcal G_2|=D-f.
\end{align*}
Then (\ref{eqn:condi-var}) becomes
\begin{align*}
    \tilde{\mathcal E}&=\mathbb E_\sigma \left[\frac{|\mathcal L_0|}{f+|\mathcal G_0|}+\frac{|\mathcal G_0|+|\mathcal L_2|}{f+|\mathcal G_0|}  \right]\\
    &=\mathbb E_\sigma \left[\frac{|\mathcal L_0|+|\mathcal G_0|+|\mathcal L_2|}{f+|\mathcal G_0|}  \right]\equiv 1.
\end{align*}
Therefore, when $f=a$, we also have $Var[\hat J_{\sigma,\pi}]=0$. \\

Next, we will consider the general case where $0<a<f\leq D$. This can be considered as a combinatorial problem where we randomly arrange $a$ type ``$O$'', $(f-a)$ type ``$\times$'' and $(D-f)$ type ``$-$'' points in a circle. We are interested in the distribution of the number of $\{O,O\}$, $\{O,\times\}$, $\{O,-\}$, $\{-,O\}$ and $\{-,\times\}$ pairs of consecutive points in clockwise direction. We consider this procedure in two steps, where we first place ``$\times$'' and ``$-$'' points, and then place ``$O$'' points.\\

\textbf{Step 1. Randomly place ``$\times$'' and ``$-$'' points on the circle.}

\vspace{0.1in}

In this step, four types of pairs may appear: $\{-,-\}$, $\{-,\times\}$, $\{\times,\times\}$ and $\{\times,-\}$. Denote $\mathcal C_1$, $\mathcal C_2$, $\mathcal C_3$ and $\mathcal C_4$ as the collections of above pairs. Since
\begin{align*}
    &|\mathcal C_1|+|\mathcal C_4|=|\mathcal C_1|+|\mathcal C_2|=D-f,\\
    &|\mathcal C_2|+|\mathcal C_3|=|\mathcal C_2|+|\mathcal C_4|=f-a,
\end{align*}
knowing the size of one set gives information on the size of all the sets. Thus, we can characterize the joint distribution by analyzing the distribution of $|\mathcal C_1|$. First, placing $(D-f)$ ``$-$'' points on a circle leads to $(D-f)$ number of $\{-,-\}$ pairs. This $(D-f)$ elements can be regarded as the borders that split the circle into $(D-f)$ bins. Now, we randomly throw $(f-a)$ number of ``$\times$'' points into these bins. If at least one ``$\times$'' falls into one bin, then the number of $\{-,-\}$ pairs ($|\mathcal C_1|$) would reduce by $1$, while $|\mathcal C_2|$ and $|\mathcal C_4|$ would increase by $1$. If $z$ ``$\times$'' points fall into one bin, then the number of $\{\times,\times\}$ ($|\mathcal C_3|$) would increase by $(z-1)$. Notice that since $s\leq D-f$ and $D-f-s\leq f-a$, we have $\max(0,D-2f+a)\leq s\leq D-f$. Consequently, for $s$ in this range, we have
\begin{align}
    P\Big\{|\mathcal C_1|=s\Big\}&=P\Big\{|\mathcal C_1|=s, |\mathcal C_3|=f-a-(D-f-s)\Big\} \nonumber\\
    &=\frac{\binom{D-f}{D-f-s}\binom{f-a-1 }{D-f-s-1}}{\binom{D-a-1}{D-f-1}} \nonumber\\
    &=\frac{\binom{D-f}{s}\binom{f-a-1}{D-f-s-1}}{\binom{D-a-1}{D-f-1}}. \label{eqn:Ps}
\end{align}
The second line is due to the stars and bars problem that the number of ways to place $n$ unlabeled balls in $m$ distinct bins such that each bin has at least one ball is $\binom{n-1}{m-1}$. For $|\mathcal C_1|=s$, we need $n=f-a$ (number of ``$\times$'') and $m=|\mathcal C_2|=D-f-s$. Moreover, the number of ways to place $n$ balls in $m$ distinct bins is $\binom{n+m-1}{m-1}$. When counting the total number of possibilities, we have $n=f-a$ and $m=D-f$. This gives the denominator. We notice that (\ref{eqn:Ps}) is actually a hyper-geometric distribution.

\newpage

\textbf{Step 2. Randomly place ``$O$'' points on the circle.}

\vspace{0.1in}

We have the probability mass function
\begin{align}
    P[\Psi]&\triangleq P\Big\{ |\mathcal L_1|=l_1, |\mathcal L_2|=l_2,|\mathcal G_0|=g_0,|\mathcal G_1|=g_1\Big\} \nonumber\\
    &=\sum_{s=D-2f+a}^{D-f-1}P\Big\{ |\mathcal L_1|=l_1, |\mathcal L_2|=l_2,|\mathcal G_0|=g_0,|\mathcal G_1|=g_1\Big| |\mathcal C_1|=s \Big\}P\Big\{ |\mathcal C_1|=s\Big\}. \label{eqn:decompose}
\end{align}
Now it remains to compute the distribution conditional on $|\mathcal C_1|$. Here we drop $|\mathcal L_0|$ because it is intrinsically determined by $|\mathcal L_1|$ and $|\mathcal L_2|$. Again, given a placement of all ``$\times$'' and ``$-$'' points, each consecutive pair can be regarded as a distinct bin. Therefore, now the problem is to randomly throw $a$ type ``$O$'' points into that $(D-a)$ bins, given that we have placed type ``$\times$'' and ``$-$'' points on the circle with $|\mathcal C_1|=s$ (and thus $|\mathcal C_2|=|\mathcal C_3|=D-f-s$ and $|\mathcal C_4|=f-a-(D-f-s)$ are also determined correspondingly). In the following, we count the number of ``$O$'' points that fall in $\mathcal C_i$, $i=1,2,3,4$, to make the event $\Psi$ happen. Note that
\begin{itemize}
    \item When at least one ``$O$'' point falls into $\mathcal C_1$ (between $\{-,-\}$), $|\mathcal L_2|$ and $|\mathcal G_0|$ increase by $1$.

    \item When at least one ``$O$'' point falls into $\mathcal C_2$ (between $\{-,\times\}$), $|\mathcal L_1|$ and $|\mathcal G_0|$ increase by $1$, while $|\mathcal G_1|$ decreases by $1$.

    \item When at least one ``$O$'' point falls into $\mathcal C_3$ (between $\{\times,-\}$), $|\mathcal L_2|$ increases by $1$.

    \item When at least one ``$O$'' point falls into $\mathcal C_4$ (between $\{\times,\times\}$), $|\mathcal L_1|$ increases by $1$.
\end{itemize}
We denote the number of bins in $\mathcal C_i$, $i=1,2,3,4$ that contain at least one ``$O$'' point as $n_1,n_2,n_3,n_4$, respectively. As a result of above reasoning, in the event $\Psi$, we have
\begin{equation*}
    \begin{cases}
        n_1+n_3=l_2,\\
        n_2+n_4=l_1,\\
        n_1+n_2=g_0,\\
        D-f-s-n_2=g_1.
    \end{cases}
\end{equation*}
Solving the equations gives
\begin{equation*}
    \begin{cases}
        n_1=g_0-(D-f-s-g_1),\\
        n_2=D-f-s-g_1,\\
        n_3=l_2-g_0+(D-f-s-g_1),\\
        n_4=l_1-(D-f-s-g_1).
    \end{cases}
\end{equation*}
Note that $\sum_{i=1}^4{n_i}=l_1+l_2$. Therefore, event $\Psi$ is equivalent to randomly pick $n_1,n_2,n_3$ and $n_4$ bins in $\mathcal C_1$,...,$\mathcal C_4$, and then distribute $a$ type ``$O$'' points in these $(l_1+l_2)$ bins such that each bin contains at least one ``$O$''. Hence, we obtain
\begin{align}
    P\Big\{ |\mathcal L_1|=l_1, |\mathcal L_2|=l_2,|\mathcal G_0|=g_0,|\mathcal G_1|=g_1\Big| |\mathcal C_1|=s \Big\}
    &=\frac{\binom{s}{n_1}\binom{D-f-s}{n_2}\binom{D-f-s}{n_3}\binom{f-a-(D-f-s)}{n_4}\binom{a-1}{l_1+l_2-1}}{\binom{D-1}{D-a-1}} \nonumber\\
    &=\frac{\binom{s}{n_1}\binom{D-f-s}{n_2}\binom{D-f-s}{n_3}\binom{f-a-(D-f-s)}{n_4}\binom{a-1}{a-l_1-l_2}}{\binom{D-1}{a}}, \label{eqn:condi_prob}
\end{align}
which is also multi-variate hyper-geometric distributed. Now combining (\ref{eqn:Ps}), (\ref{eqn:decompose}) and (\ref{eqn:condi_prob}), we obtain the joint distribution of $|\mathcal L_0|,|\mathcal L_1|,|\mathcal L_2|,|\mathcal G_0|$ and $|\mathcal G_1|$ as
\begin{align}
    &P\Big\{ |\mathcal L_1|=l_1, |\mathcal L_2|=l_2,|\mathcal G_0|=g_0,|\mathcal G_1|=g_1\Big\} \nonumber\\
    &=\sum_{s=\max(0,D-2f+a)}^{D-f-1}\frac{\binom{s}{n_1}\binom{D-f-s}{n_2}\binom{D-f-s}{n_3}\binom{f-a-(D-f-s)}{n_4}\binom{a-1}{a-l_1-l_2}}{\binom{D-1}{a}}\cdot \frac{\binom{D-f}{s}\binom{f-a-1}{D-f-s-1}}{\binom{D-a-1}{D-f-1}}. \label{eqn:density}
\end{align}
Now let $\Xi$ be the feasible set of $(l_0,l_1,g_0,g_1,g_2)$ that satisfies the intrinsic constraints (\ref{eqn:constraint append}). The desired expectation w.r.t. both $\pi$ and $\sigma$ can thus be written as
\begin{align*}
    \tilde{\mathcal E}&=\sum_{\Xi}\left(\frac{l_0}{f+g_0+g_1}+\frac{a(g_0+l_2)}{(f+g_0+g_1)f}\right)\cdot\\
    &\hspace{0.2in} \left(\sum_{s=\max(0,D-2f+a)}^{D-f-1}\frac{\binom{s}{n_1}\binom{D-f-s}{n_2}\binom{D-f-s}{n_3}\binom{f-a-(D-f-s)}{n_4}\binom{a-1}{a-l_1-l_2}}{\binom{D-1}{a}}\cdot \frac{\binom{D-f}{s}\binom{f-a-1}{D-f-s-1}}{\binom{D-a-1}{D-f-1}}\right).
\end{align*}
The desired result can then follows by (\ref{eqn:var-formula}).
\end{proof}

\subsection{Proof of Proposition \ref{prop:symmetry}} \label{sec:prop:symmetry proof}

\begin{manualproposition}{3.2}[Symmetry]
$Var[\hat J_{\sigma,\pi}]$ is the same for the $(D,f,a)$-data pair and the $(D,f,f-a)$-data pair, $\forall 0\leq a\leq f\leq D$.
\end{manualproposition}

\begin{proof}
For fixed $a,f,D$, let $\tilde{\mathcal E}_1$ be the expectation defined in Theorem~\ref{theo:CMH-sigma,pi var} for $(\bm v_1,\bm w_1)$, and $\tilde{\mathcal E}_2$ be that for $(\bm v_2,\bm w_2)$. From Theorem~\ref{theo:CMH-sigma,pi var} we know that
\begin{align*}
    \tilde{\mathcal E}_1=\mathbb E_{(l_0,l_2,g_0,g_1)}\Big[\frac{l_0}{f+g_0+g_1}+\frac{a(g_0+l_2)}{(f+g_0+g_1)f}\Big],
\end{align*}
where $(l_0,l_2,g_0,g_1)$ follows the distribution of $(|\mathcal L_0|,|\mathcal L_2|,|\mathcal G_0|,|\mathcal G_1|)$ associated with the location vector $\bm x_1$ of $(\bm v_1,\bm v_2)$. For data pair $(\bm v_2,\bm w_2)$, we can consider its location vector $\bm x_2$ as swapping the ``$O$'' and ``$\times$'' entries of $\bm x_1$. Now we denote the size of the corresponding sets (Definition \ref{def-sets}) of $\bm x_2$ as $l_i's,g_i's,h_i's$, for $i=0,1,2$. Since $\sigma$ is applied before hashing, by symmetry there is a one-to-one correspondence between the two location vectors. More specifically, $l_0'$ corresponds to $h_1$, $g_0'$ corresponds to $g_1$, $g_1'$ corresponds to $g_0$, and $l_2'$ corresponds to $h_2$. Therefore, in probability we can write

\begin{align*}
    \tilde{\mathcal E}_2&=\mathbb E_{(l_0',l_2',g_0',g_1')}\Big[\frac{l_0'}{f+g_0'+g_1'}+\frac{a(g_0'+l_2')}{(f+g_0'+g_1')f}\Big]\\
    &=\mathbb E_{(h_1,h_2,g_0,g_1)}\Big[\frac{h_1}{f+g_0+g_1}+\frac{(f-a)(g_1+h_2)}{(f+g_0+g_1)f}\Big].
\end{align*}

Consequently, we have
\begin{align*}
    \tilde{\mathcal E}_1-\tilde{\mathcal E}_2=\mathbb E_{(l_0,l_2,h_1,h_2,g_0,g_1)}\Big[\frac{l_0-h_1}{f+g_0+g_1}+\frac{a(g_0+l_2)-(f-a)(g_1+h_2)}{(f+g_0+g_1)f}\Big].
\end{align*}
In the sequel, the subscript of expectation is suppressed for conciseness. Exploiting the constraints (\ref{eqn:constraint append}), we deduce that $h_1=(f-a)-l_1-g_1$,  $h_2=l_0+g_0+l_1+g_1-a$ and $l_0+l_1=a-l_2$. Using these facts we obtain
\begin{align*}
    \tilde{\mathcal E}_1-\tilde{\mathcal E}_2&=\mathbb E\Big[\frac{(l_0-(f-a)+l_1+g_1)f+a(g_0+l_2)-(f-a)(l_0+g_0+l_1+2g_1-a)}{(f+g_0+g_1)f}\Big]\\
    &=\mathbb E\Big[\frac{(2a-f+g_1-l_2)f+a(g_0+l_2)-(f-a)(2g_1+g_0-l_2)}{(f+g_0+g_1)f}\Big]\\
    &=\mathbb E\Big[\frac{2(f+g_0+g_1)a-(f+g_0+g_1)f}{(f+g_0+g_1)f}\Big]\\
    &=2J-1.
\end{align*}
Comparing the variances of $\hat J_{\sigma,\pi}(\bm v_1, \bm w_1)$ and $\hat J_{\sigma,\pi}(\bm v_2, \bm w_2)$, we derive
\begin{align*}
    &Var[\hat J_{\sigma,\pi}(\bm v_1, \bm w_1)]-Var[\hat J_{\sigma,\pi}(\bm v_2, \bm w_2)]\\
    &=(\frac{J}{K}+\frac{(K-1)\tilde{\mathcal E_1}}{K}-J^2)-(\frac{1-J}{K}+\frac{(K-1)\tilde{\mathcal E_2}}{K}-(1-J)^2)\\
    &=-\frac{K-1}{K}(2J-1)+\frac{K-1}{K}(\tilde{\mathcal E_1}-\tilde{\mathcal E_2})=0.
\end{align*}
This completes the proof.
\end{proof}

\subsection{Proof of Lemma \ref{lemma:increment}} \label{sec:lemma:increment proof}

\begin{manuallemma}{3.3}[Increasing Increment]
Assume $a>0$ and $f>a$ are arbitrary and fixed. Denote $\tilde{\mathcal E}_D$ as in \eqref{eqn:E app} in Theorem~\ref{theo:CMH-sigma,pi var}, with $D$ treated as a parameter. Then we have $\tilde{\mathcal E}_{D+1}> \tilde{\mathcal E}_D$ for $\forall D\geq f$.
\end{manuallemma}

\begin{proof}
Let the probability mass function (\ref{eqn:density}) with $a$, $f$ and dimension $D$ be $P_{a,f,D}(l_0,l_2,g_0,g_1)$. Conditional on $l_0,l_2,g_0,g_1$ with $D$ elements, the possible values $l_0',l_2',g_0',g_1'$ when adding a ``$-$'' are
\begin{itemize}
    \item $g_0'=g_0+1,l_0'=l_0,l_2'=l_2,g_1'=g_1$. This is true when the new elements falls between a pair of $(\times,O)$, with probability $\frac{l_1+l_2-g_0}{D}$.

    \item $g_1'=g_1+1,l_0'=l_0,l_2'=l_2,g_0'=g_0$, when the new elements falls between a pair of $(\times,\times)$, with probability $\frac{f-a-l_1-g_1}{D}$.

    \item $g_1'=g_1+1,l_2'=l_2+1,l_0'=l_0,g_0'=g_0$, when the new elements falls between a pair of $(O,\times)$, with probability $\frac{l_1}{D}$.

    \item $l_0'=l_0-1,l_2'=l_2+1,g_0'=g_0+1,g_1'=g_1$, when the new elements falls between a pair of $(O,O)$, with probability $\frac{l_0}{D}$.

    \item All values unchanged, when the ``$-$'' falls between other types of pairs, with probability $\frac{D-f+g_0+g_1}{D}$.
\end{itemize}
Denote $\Xi_D$ as the feasible set satisfying (\ref{eqn:constraint append}) with dimension $D\geq f$. Above reasoning builds a correspondence between $\Xi_D$ and $\Xi_{D+1}$. More precisely, we have
\begin{align*}
    \tilde{\mathcal E}_{D+1}&=\sum_{\Xi_{D+1}}\left(\frac{l_0'}{f+g_0'+g_1'}+\frac{a(g_0'+l_2')}{(f+g_0'+g_1')f}\right)P_{a,f,D+1}(l_0',l_2',g_0',g_1')\\
    &=\sum_{\Xi_{D}}\bigg\{\Big(\frac{l_0}{f+g_0+g_1+1}+\frac{a(g_0+l_2+1)}{(f+g_0+g_1+1)f}\Big)\frac{l_1+l_2-g_0}{D}P_{a,f,D}(l_0,l_2,g_0,g_1)\\
    &\hspace{0.3in}+\Big(\frac{l_0}{f+g_0+g_1+1}+\frac{a(g_0+l_2)}{(f+g_0+g_1+1)f}\Big)\frac{f-a-l_1-g_1}{D}P_{a,f,D}(l_0,l_2,g_0,g_1)\\
    &\hspace{0.3in}+\Big(\frac{l_0}{f+g_0+g_1+1}+\frac{a(g_0+l_2+1)}{(f+g_0+g_1+1)f}\Big)\frac{l_1}{D}P_{a,f,D}(l_0,l_2,g_0,g_1)\\
    &\hspace{0.3in}+\Big(\frac{l_0-1}{f+g_0+g_1+1}+\frac{a(g_0+l_2+2)}{(f+g_0+g_1+1)f}\Big)\frac{l_0}{D}P_{a,f,D}(l_0,l_2,g_0,g_1)\\
    &\hspace{0.3in}+\Big(\frac{l_0}{f+g_0+g_1}+\frac{a(g_0+l_2)}{(f+g_0+g_1)f}\Big)\frac{D-f+g_0+g_1}{D}P_{a,f,D}(l_0,l_2,g_0,g_1)\bigg\}.\\
\end{align*}
The increment can be computed as
\begin{align}
    \tilde\delta_D&\triangleq \tilde{\mathcal E}_{D+1}-\tilde{\mathcal E}_{D} \nonumber\\
    &=\sum_{\Xi_D}\bigg\{\frac{f-g_0-g_1}{D}\Big[ \big(\frac{l_0}{f+g_0+g_1+1}-\frac{l_0}{f+g_0+g_1}\big)+\big(\frac{a(g_0+l_2+1)}{f+g_0+g_1+1}-\frac{a(g_0+l_2)}{f+g_0+g_1}\big) \Big] \nonumber\\
    &\hspace{0.5in} -\frac{l_0}{D(f+g_0+g_1+1)}-\frac{a(f-a-l_1-g_1)-al_0}{Df(f+g_0+g_1+1)}\bigg\}P_{a,f,D}(l_0,l_2,g_0,g_1) \nonumber\\
    &=\sum_{\Xi_D}\bigg\{\frac{(f-g_0-g_1)[a(f+g_1-l_2)-fl_0]}{Df(f+g_0+g_1)(f+g_0+g_1+1)}-\frac{(f-a)l_0+a(f-a-l_1-g_1)}{Df(f+g_0+g_1+1)}\bigg\}P_{a,f,D}(l_0,l_2,g_0,g_1) \nonumber\\
    &=\sum_{\Xi_D}\frac{2af(l_1+g_1)-2f(f-a)l_0-2a(f-a)(g_0+g_1)}{Df(f+g_0+g_1)(f+g_0+g_1+1)}P_{a,f,D}(l_0,l_2,g_0,g_1) \nonumber\\
    &=\mathbb E\Big[\frac{2af(l_1+g_1)-2f(f-a)l_0-2a(f-a)(g_0+g_1)}{Df(f+g_0+g_1)(f+g_0+g_1+1)}\Big] \nonumber\\
    &=\mathbb E\Big[\frac{2af(f-a-h_1)-2f(f-a)l_0-2a(f-a)(g_0+g_1+f-f)}{Df(f+g_0+g_1)(f+g_0+g_1+1)}\Big] \nonumber\\
    &=\mathbb E\Big[\frac{4a(f-a)}{D(f+g_0+g_1)(f+g_0+g_1+1)}\Big]-\mathbb E\Big[\frac{2ah_1+2(f-a)l_0}{D(f+g_0+g_1)(f+g_0+g_1+1)}\Big]-\mathbb E\Big[\frac{2a(f-a)}{Df(f+g_0+g_1+1)}\Big] \nonumber\\
    &\triangleq 4a(f-a)E_0-2aE_1-2(f-a)E_2-2a(f-a)E_3, \label{lemma2-eqn:increment}
\end{align}
where
\begin{align*}
    E_0&=\mathbb E\Big[\frac{1}{D(f+g_0+g_1)(f+g_0+g_1+1)}\Big],\ \ E_1=\mathbb E\Big[\frac{h_1}{D(f+g_0+g_1)(f+g_0+g_1+1)}\Big],\\
    E_2&=\mathbb E\Big[\frac{l_0}{D(f+g_0+g_1)(f+g_0+g_1+1)}\Big],\ \
    E_3=\mathbb E\Big[\frac{g_2}{Df(f+g_0+g_1+1)}\Big].
\end{align*}
Note that here the expectations are taken w.r.t. the set size distribution with $a,f,D$. We can expand the terms of density function (\ref{eqn:density}) to derive
\begin{align*}
    &P_{a,f,D}(l_0,l_2,g_0,g_1) \\
    &=\sum_{s=\max(0,D-2f+a)}^{D-f-1}\frac{(D-f-s)(D-f)!(f-a-1)!}{[D-(f+g_0+g_1)]![(f+g_0+g_1)-D+s]!g_1!(D-f-s-g_1)!}\\
    &\frac{(a-1)!}{(g_0+g_1-l_2)![D-s+l_2-(f+g_0+g_1)]!(f-a-l_1-g_1)!(f+g_1+l_1-D+s)!l_0!(a-l_0-1)!}\\
    &\frac{a!(f-a)!(D-f-1)!}{(D-1)!}.
\end{align*}
Denote $a'=a-1$, $f'=f-1$, $D'=D-1$ and $l_0'=l_0-1$. We have
\begin{align*}
    E_2&=\sum_{\Xi_D}\frac{l_0}{D(f+g_0+g_1)(f+g_0+g_1+1)}P_{a,f,D}(l_0,l_2,g_0,g_1)\\
    &=\sum_{\Xi_D}\frac{a(a-1)}{D-1}\cdot\frac{1}{D(f+g_0+g_1)(f+g_0+g_1+1)}\\
    &\hspace{0.3in} \sum_{s=\max(0,D'-2f'+a')}^{D'-f'-1}\frac{(D'-f'-s)(D'-f')!(f'-a'-1)!}{[D'-(f'+g_0+g_1)]![(f'+g_0+g_1)-D'+s]!g_1!(D'-f'-s-g_1)!}\\
    &\frac{(a'-1)!}{(g_0+g_1-l_2)![D'-s+l_2-(f'+g_0+g_1)]!(f'-a'-l_1-g_1)!(f'+g_1+l_1-D'+s)!l_0'!(a'-l_0'-1)!}\\
    &\frac{a'!(f'-a')!(D'-f'-1)!}{(D'-1)!}\\
    &=\sum_{\Xi_{D-1}}\frac{a(a-1)}{D-1}\frac{1}{D(f+g_0+g_1)(f+g_0+g_1+1)}P_{a-1,f-1,D-1}(l_0,l_2,g_0,g_1)\\
    &=\frac{a(a-1)}{D-1}\mathbb E_{a-1,f-1,D-1}\Big[\frac{1}{D(f+g_0+g_1)(f+g_0+g_1+1)}\Big]\\
    &\triangleq \frac{a(a-1)}{D-1}\bar E.
\end{align*}
Here the subscript means that we are taking expectation w.r.t the set sizes when the number of ``$O$'', ``$\times$'' and ``$-$'' points is $(a-1,f-1,D-1)$. By symmetry, it can be shown similarly that
\begin{align*}
    E_1&=\frac{(f-a)(f-a-1)}{D-1}\mathbb E_{a,f-1,D-1}\Big[\frac{1}{D(f+g_0+g_1)(f+g_0+g_1+1)}\Big]= \frac{(f-a)(f-a-1)}{D-1} \bar E.
\end{align*}
Substituting above results into (\ref{lemma2-eqn:increment}), we obtain
\begin{align*}
    \tilde\delta_D&=2a(f-a)[2E_0-\frac{f-2}{D-1}\bar E-E_3].
\end{align*}
To compute $E_0$, note that with $a,f$ and $D$, variable $g_2$ is distributed as hyper($D-1,D-f,D-f-1$). For $\bar E$, the distribution becomes hyper($D-2,D-f,D-f-1$). Since $f+g_0+g_1=D-g_2$, we deduce
\begin{align*}
    E_0&=\sum_{s=\max(0,D-2f)}^{D-f-1}\frac{1}{D(D-s)(D-s+1)}\frac{\binom{D-f-1}{s}\binom{f}{D-f-s}}{\binom{D-1}{D-f}}\\
    &=\sum_{s=\max(0,D-2f)}^{D-f-1}\frac{1}{D(D-s)(D-s+1)}\frac{(D-f-1)!f!}{s!(D-f-s-1)!(D-f-s)!(-D+2f+s)!}\frac{(D-f)!(f-1)!}{(D-1)!},
\end{align*}
and
\begin{align*}
    \bar E&=\sum_{s=\max(0,D-2f+1)}^{D-f-1}\frac{1}{D(D-s)(D-s+1)}\frac{\binom{D-f-1}{s}\binom{f-1}{D-f-s}}{\binom{D-2}{D-f}}\\
    &=\sum_{s=\max(0,D-2f+1)}^{D-f-1}\frac{1}{D(D-s)(D-s+1)}\frac{(D-f)!(f-2)!}{(D-2)!}\cdot\\
    &\hspace{2in}\frac{(D-f-1)!(f-1)!}{s!(D-f-s-1)!(D-f-s)!(-D+2f+s-1)!}.
\end{align*}
For $\forall D\geq f$, we have
\begin{align*}
    \frac{f-2}{D-1}\bar E&\leq \sum_{s=\max(0,D-2f)}^{D-f-1}\frac{(f-2)(D-1)(-D+2f+s)}{D(D-1)f(f-1)(D-s)(D-s+1)}\\
    &\hspace{0.6in} \frac{(D-f-1)!f!}{s!(D-f-s-1)!(D-f-s)!(-D+2f+s)!}\frac{(D-f)!(f-1)!}{(D-1)!}\\
    &\leq \mathbb E\Big[\frac{(f-2)(f-(D-f-g_2))}{Df(f-1)(D-g_2)(D-g_2+1)}\Big]\\
    &< \mathbb E\Big[\frac{(f-g_0-g_1)}{Df(f+g_0+g_1)(f+g_0+g_1+1)}\Big].
\end{align*}
Consequently, we have
\begin{align*}
    \tilde\delta_D&> 2a(f-a)\mathbb E\Big[\frac{2}{D(f+g_0+g_1)(f+g_0+g_1+1)}-\frac{f-g_0-g_1}{Df(f+g_0+g_1)(f+g_0+g_1+1)}\\
    &\hspace{2in}-\frac{f+g_0+g_1}{Df(f+g_0+g_1)(f+g_0+g_1+1)}\Big]\\
    &= 0,
\end{align*}
and note that this holds for $\forall D\geq K$. The proof is complete.
\end{proof}

\subsection{Proof of Theorem \ref{theo:smaller-variance}}  \label{sec:theo:smaller-variance proof}

\begin{manualtheorem}{3.4}[Uniform Superiority]
For any two binary vectors $\bm v,\bm w\in\{0,1\}^D$ with $J\neq 0$ or $1$, it holds that $Var[\hat J_{\sigma,\pi}(\bm v,\bm w)]< Var[\hat J_{MH}(\bm v,\bm w)]$.
\end{manualtheorem}

\begin{proof}
By assumption we have $0<a<f$. To compare $Var[\hat J_{\sigma,\pi}]$ with $Var[\hat J_{MH}]=\frac{J(1-J)}{K}=\frac{J}{K}+\frac{(K-1)J^2}{K}-J^2$, it suffices to compare $\tilde{\mathcal E}$ with $J^2$. When $D=f$, we know that the location vector $\bm x$ of $(\bm v,\bm w)$ contains no ``$-$'' elements. It is easy to verify that in this case, $|\mathcal G_0|=|\mathcal G_1|=|\mathcal L_2|=0$, and $|\mathcal L_0|$ follows hyper($f-1,a,a-1$). By Theorem~\ref{theo:CMH-sigma,pi var}, it follows that when $D=f$,
\begin{align*}
    \tilde{\mathcal E}_D&=\frac{1}{f}\mathbb E[|\mathcal L_0|]=\frac{a(a-1)}{f(f-1)}=J\tilde J<J^2.
\end{align*}
Recall the definition $\tilde J=\frac{a-1}{f-1}$, which is always less than $J$. On the other hand, as $D\rightarrow \infty$, we have $|\mathcal L_0|\rightarrow 0$, $|\mathcal L_2|\rightarrow a$, $|\mathcal G_0|\rightarrow a$ and $|\mathcal G_1|\rightarrow f-a$. We can easily show that
\begin{align*}
    \tilde{\mathcal E}_D\rightarrow J^2,\hspace{0.1in} \text{as}\ D\rightarrow\infty.
\end{align*}
By Lemma~\ref{lemma:increment}, the sequence $(\tilde{\mathcal E}_f,\tilde{\mathcal E}_{f+1},\tilde{\mathcal E}_{f+2},...)$ is strictly increasing. Since it is convergent with limit $J^2$, by the Monotone Convergence Theorem we know that $\tilde{\mathcal E_D}<J^2$, $\forall D\geq f$.
\end{proof}

\subsection{Proof of Proposition \ref{prop:same-ratio}}  \label{sec:prop:same-ratio proof}

\begin{manualproposition}{3.5}[Consistent Improvement]
Suppose $f$ is fixed. In terms of  $a$, the variance ratio $\rho(a)=\frac{Var[\hat J_{MH}(\bm v,\bm w)]}{Var[\hat J_{\sigma,\pi}(\bm v,\bm w)]}$ is constant for any $0<a<f$.
\end{manualproposition}

\begin{proof}
Let $\tilde{\mathcal E}$ be defined as in Theorem~\ref{theo:CMH-sigma,pi var}. Assume that $D$ and $f$ are fixed and $a$ is variable. Firstly, we can write the variance ratio explicitly as
\begin{align*}
    \rho(a)&=\frac{\frac{J-J^2}{K}}{\frac{J}{K}+\frac{(K+1)\tilde{\mathcal E}}{K}-J^2}
    =\frac{1-J}{1-J-(K-1)(J-\frac{\tilde{\mathcal E}}{J})}.
\end{align*}
We now show that the term $J-\frac{\tilde{\mathcal E}}{J}=C(1-J)$, where $C$ is some constant independent of $J$ (i.e., $a$). Then, for fixed $D$ and $f$, by cancellation $\rho(a)$ would be constant for all $0<a<f$. We have
\begin{align}
    J-\frac{\tilde{\mathcal E}}{J}&=\frac{a}{f}-\mathbb E_{a,f,D}\Big[\frac{fl_0}{a(f+g_0+g_1)}+\frac{g_0+l_2}{f+g_0+g_1}\Big] \nonumber\\
    &=\mathbb E\Big[\frac{a^2(f+g_0+g_1)-f^2l_0-af(g_0+l_2)}{af(f+g_0+g_1)}\Big] \nonumber\\
    &=\mathbb E\Big[\frac{a(a-f)(g_0+g_1)+a^2f+afg_1-f^2l_0-afl_2}{af(f+g_0+g_1)}\Big] \nonumber\\
    &=\mathbb E\Big[\frac{a(a-f)(g_0+g_1)+af(l_0+l_1)+afg_1-f^2l_0}{af(f+g_0+g_1)}\Big] \nonumber\\
    &=\mathbb E\Big[\frac{a(a-f)(g_0+g_1)+f(a-f)l_0+af(f-a-h_1)}{af(f+g_0+g_1)}\Big], \label{prop2-eqn1}
\end{align}
where we use the constraints (\ref{eqn:constraint append}) that $l_0+l_1+l_2=a$ and $l_1+g_1+h_1=f-a$. We now study the three terms respectively. We have
\begin{align*}
    \mathbb E\Big[\frac{a(a-f)(g_0+g_1)}{af(f+g_0+g_1)}\Big]=-(1-J)\mathbb E\Big[\frac{g_0+g_1}{f+g_0+g_1}\Big]\triangleq -E'(1-J).
\end{align*}
We have shown in the proof of Lemma~\ref{lemma:increment} that
\begin{align*}
    \mathbb E_{a,f,D}\Big[\frac{l_0}{f+g_0+g_1}\Big]=\frac{a(a-1)}{D-1}\mathbb E_{a-1,f-1,D-1}\Big[\frac{1}{f+g_0+g_1}\Big]\triangleq \frac{a(a-1)}{D-1}E^*,
\end{align*}
and by symmetry it holds that
\begin{align*}
    \mathbb E_{a,f,D}\Big[\frac{h_1}{f+g_0+g_1}\Big]= \frac{(f-a)(f-a-1)}{D-1}E^*.
\end{align*}
Note that Since $f$ is fixed, $(|\mathcal G_0|+|\mathcal G_1|)$ is distributed independent of $a$. Consequently, $E'$ and $E^*$ are both independent of $a$. Next, we obtain
\begin{align*}
    \mathbb E\Big[\frac{f(a-f)l_0}{af(f+g_0+g_1)}\Big]=-(1-J)\frac{f(a-1)}{D-1}E^*,
\end{align*}
and
\begin{align*}
    \mathbb E\Big[\frac{af(f-a-h_1)}{af(f+g_0+g_1)}\Big]&=(1-J)fE^*-(1-J)\frac{f(f-a-1)}{D-1}E^*.
\end{align*}
Summing up the terms and substituting into (\ref{prop2-eqn1}), we derive
\begin{align*}
    J-\frac{\tilde{\mathcal E}}{J}&=C(1-J),
\end{align*}
where $C=-E'+(f-\frac{f(f-2)}{D-1})E^*$, which is independent of $a$. Taking into $\rho(a)$, we get
\begin{align*}
    \rho(a)=\frac{1-J}{1-J-(K-1)C(1-J)}=\frac{1}{1-(K-1)C},
\end{align*}
which is a constant only depending on $f$, $D$ and $K$. This completes the proof.
\end{proof}

\vspace{0.5in}

\bibliography{standard}
\bibliographystyle{plainnat}

\end{document}